\documentclass[10pt,journal,compsoc]{IEEEtran}

\ifCLASSOPTIONcompsoc
  \usepackage[nocompress]{cite}
\else
  \usepackage{cite}
\fi

\usepackage{graphicx}
\usepackage{multirow}
\usepackage{array}
\usepackage{booktabs}
\usepackage{amsmath,amssymb}
\usepackage{ragged2e}
\usepackage{subfigure}

\usepackage{algorithm}
\usepackage{algorithmic}

\usepackage[caption=false,font=footnotesize,labelfont=sf,textfont=sf]{subfig}
\usepackage{float}

\ifCLASSINFOpdf
\else
\fi

\begin{document}
\title{Joint Face Alignment and 3D Face Reconstruction with Application to Face Recognition}
\author{
Feng Liu,~\IEEEmembership{Member,~IEEE},
Qijun~Zhao,~\IEEEmembership{Member,~IEEE},
		Xiaoming Liu,~\IEEEmembership{Member,~IEEE}
        and~Dan Zeng~\IEEEmembership{}% <-this % stops a space
\IEEEcompsocitemizethanks{
%\IEEEcompsocthanksitem A preliminary version of this work \cite{liu2016joint} has been presented on the 14th European Conference on Computer Vision (ECCV2016).
%of Electrical and Computer Engineering, Georgia Institute of Technology, Atlanta,
\IEEEcompsocthanksitem Feng Liu, Dan Zeng and Qijun Zhao are with the National Key Laboratory of Fundamental Science on Synthetic Vision, College of Computer Science, Sichuan University, Chengdu, Sichuan 610065, P. R. China. Qijun Zhao is the corresponding author, reachable at qjzhao@scu.edu.cn. 

%Xiaoming Liu is with the Dept. of Computer Science and Engineering, Michigan State University, East Lansing, MI 48824, U.S.A.
\IEEEcompsocthanksitem Xiaoming Liu is with the Dept. of Computer Science and Engineering, Michigan State University, East Lansing, MI 48824, U.S.A.
\protect\\}
%E-mail: see http://www.michaelshell.org/contact.html
%\IEEEcompsocthanksitem Feng Liu, Dan Zeng and Qijun Zhao are with the College of Computer Science, Sichuan University. Xiaoming Liu is with the Department of Computer Science and Engineering, Michigan State University.}% <-this % stops an unwanted space
}

%\markboth{Journal of \LaTeX\ Class Files,~Vol.~14, No.~8, December~2016}%
%{Shell \MakeLowercase{\textit{et al.}}: Bare Demo of IEEEtran.cls for Computer Society Journals}

\IEEEtitleabstractindextext{%
\justifying  
\begin{abstract}
Face alignment and 3D face reconstruction are traditionally accomplished as separated tasks. By exploring the strong correlation between 2D landmarks and 3D shapes, in contrast, we propose a joint face alignment and 3D face reconstruction method to simultaneously solve these two problems for 2D face images of arbitrary poses and expressions. This method, based on a summation model of 3D faces and cascaded regression in 2D and 3D shape spaces, iteratively and alternately applies two cascaded regressors, one for updating 2D landmarks and the other for 3D shape. The 3D shape and the landmarks are correlated via a 3D-to-2D mapping matrix, which is updated in each iteration to refine the location and visibility of 2D landmarks. Unlike existing methods, the proposed method can fully automatically generate both pose-and-expression-normalized (PEN) and expressive 3D faces and localize both visible and invisible 2D landmarks. Based on the PEN 3D faces, we devise a method to enhance face recognition accuracy across poses and expressions. Both linear and nonlinear implementations of the proposed method are presented and evaluated in this paper. Extensive experiments show that the proposed method can achieve the state-of-the-art accuracy in both face alignment and 3D face reconstruction, and benefit face recognition owing to its reconstructed PEN 3D face.

\end{abstract}

}

\maketitle
\IEEEdisplaynontitleabstractindextext
\IEEEpeerreviewmaketitle

\IEEEraisesectionheading{\section{Introduction}\label{sec:introduction}}
\IEEEPARstart{T}HREE-dimensional (3D) face models have recently been employed to assist pose or expression invariant face recognition and achieve state-of-the-art performance \cite{blanz2003face, han20123d, zhu2015high}. A crucial step in these 3D face assisted face recognition methods is to reconstruct the 3D face model from a two-dimensional (2D) face image. Besides its applications in face recognition, 3D face reconstruction is also useful in other face-related tasks, e.g., facial expression analysis~\cite{chu20143d, yin20063d} and facial animation~\cite{cao20133d, cao2016real}. While many 3D face reconstruction methods are available, they mostly require landmarks on the face image as input, and are difficult to handle large-pose faces that have invisible landmarks due to self-occlusion.

\begin{figure}[!htb]
\begin{center}
\includegraphics[width=3.4in]{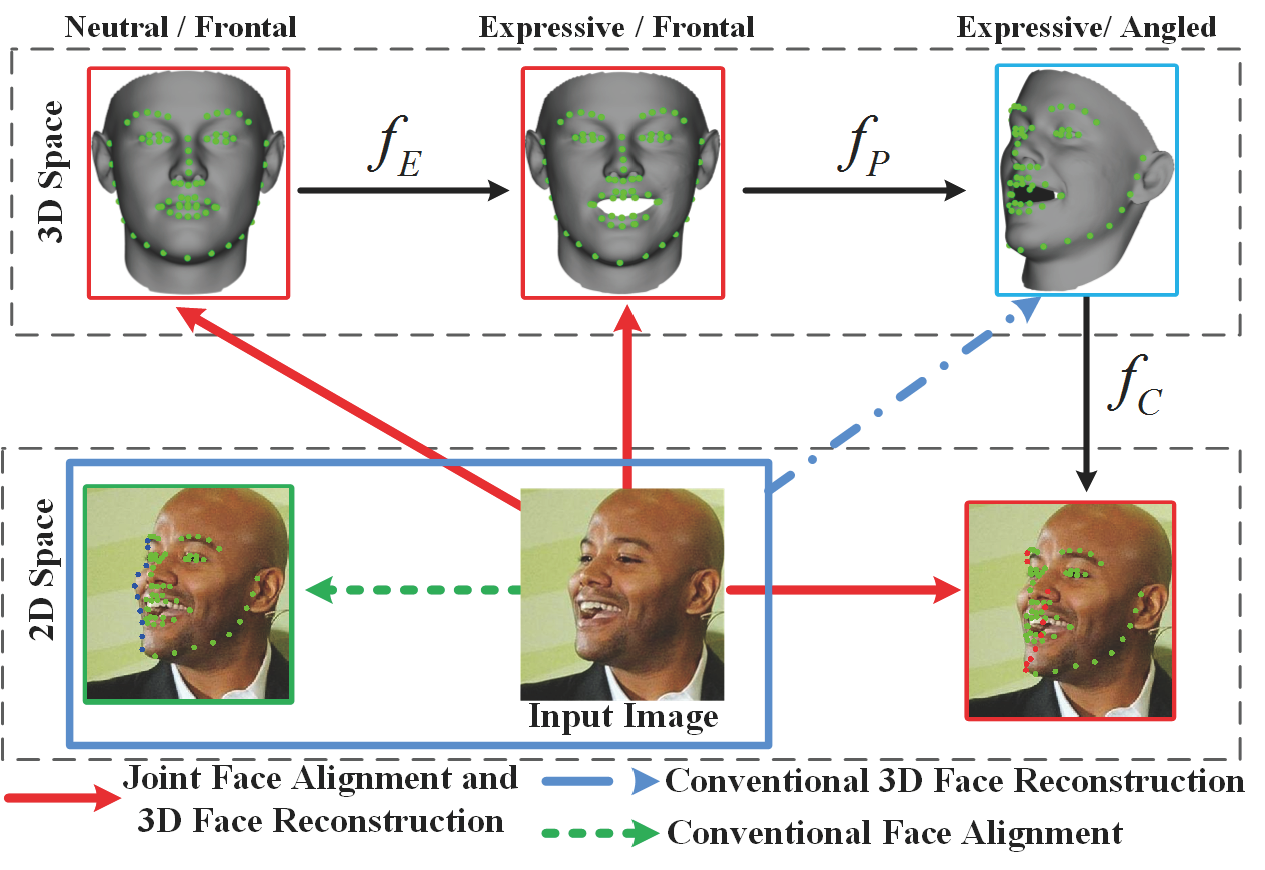}
\end{center}
   \caption{We view 2D landmarks are generated from a 3D face through 3D expression ($f_E$) and pose ($f_P$) deformation, and camera projection ($f_C$). While conventional face alignment and 3D face reconstruction are two {\it separated} tasks and the latter requires the former as input, this paper performs these two tasks {\it jointly}, i.e., reconstructing a 3D face and estimating visible/invisible landmarks (green/red points) from a 2D face image with arbitrary poses and expressions.}
\label{fig:problem}
\end{figure}

Existing studies tackle the problems of facial landmark localization (or face alignment) and 3D face reconstruction \emph{separately}. However, these two problems are chicken-and-egg problems. On one hand, 2D face images are projections of 3D faces onto the 2D plane. Given a 3D face and a 3D-to-2D mapping function, it is easy to compute the visibility and position of 2D landmarks. On the other hand, the landmarks provide rich information about facial geometry, which is the basis of 3D face reconstruction. Figure \ref{fig:problem} illustrates the relationship between 2D landmarks and 3D faces. That is, the visibility and position of landmarks in the projected 2D image are determined by four factors: the 3D shape, the deformation due to expression and pose, and the camera projection parameters. \emph{Given such a clear correlation between 2D landmarks and 3D shape, it is evident that ideally they should be solved jointly, instead of separately as in prior works - indeed this is the core of this work.}

Motivated by the aforementioned observation, this paper proposes a unified framework to simultaneously solve the two problems of face alignment and 3D face reconstruction. Two sets of regressors are jointly learned from a training set of pairing annotated 2D face images and 3D face shapes. Based on the texture features around landmarks on a face image, one set of regressors (called landmark regressors) gradually move the landmarks towards their true positions. By utilizing the facial landmarks as clues, the other set of regressors (called shape regressors) gradually improve the reconstructed 3D face. These two sets of regressors are alternately and iteratively applied. Specifically, in each iteration, adjustment to the landmarks is firstly estimated via the landmark regressors, and this landmark adjustment is also used to estimate 3D shape adjustment via the shape regressors. The 3D-to-2D mapping is then computed based on the adjusted 3D shape and 2D landmarks, and it further refines the landmarks.

A preliminary version of this work was published in the 14th European Conference on Computer Vision (ECCV) 2016~\cite{liu2016joint}. We further extend the work from four aspects. (i) We explicitly reconstruct expression deformation of 3D faces, so that both PEN (pose and expression normalized) and expressive 3D faces can be reconstructed. (ii) We implement the proposed method in both linear and nonlinear regressions. (iii) We present in detail the application of the proposed method to face recognition. (iv) We carry out a more extensive evaluation with comparisons to state-of-the-art methods. In summary, this paper makes the following contributions.
\begin{itemize}
\item We present a novel cascaded coupled-regressor based method with linear and non-linear regressions for joint face alignment and 3D face reconstruction from a single 2D image of arbitrary pose and expression.
\item By integrating 3D shape information, the proposed method can more accurately localize landmarks on images of arbitrary view angles in [-$90^{\circ}, 90^{\circ}$].
\item We explicitly deal with expression deformation of 3D faces, so that both PEN and expressive 3D faces can be reconstructed at a high accuracy.
\item We propose a 3D-enhanced approach to improve face recognition accuracy on off-angle and expressive face images based on the reconstructed PEN 3D faces.
\item We achieve state-of-the-art 3D face reconstruction and face alignment performance on BU3DFE~\cite{yin20063d}, AFLW~\cite{K2011Annotated}, and AFLW2000 3D~\cite{zhu2016CVPR} databases. We investigate the other-race effect on 3D reconstruction of the proposed method on FRGC v2.0 database~\cite{phillips2005overview}. We demonstrate the effectiveness of our proposed 3D-enhanced face recognition method in improving state-of-the-art deep learning based face matchers on Multi-PIE~\cite{gross2010multi} and CFP~\cite{sengupta2016frontal} databases.
\end{itemize}

The rest of this paper is organized as follows. Section~\ref{sec:prior} briefly reviews related work in the literature. Section~\ref{sec:method} introduces in detail the proposed joint face alignment and 3D face reconstruction method and two alternative implementations. Section~\ref{sec:facerecognition} shows its application to face recognition. Section~\ref{sec:exp} reports the experimental results. Section~\ref{sec:con}  concludes the paper.

\section{Prior Work}
\label{sec:prior}
\subsection{Face Alignment}
Classical face alignment methods, e.g., Active Shape Model (ASM)~\cite{tecootes1994active, cristinacce2007boosted} or Active Appearance Model (AAM)~\cite{cootes2001active, matthews2004active,face-model-fitting-on-low-resolution-images, discriminative-face-alignment}, search for landmarks based on global shape models and texture models. Constrained Local Model (CLM)~\cite{cristinacce2008automatic} also utilizes global shape models to regularize the landmark locations, but it employs discriminative local texture models. Regression based methods~\cite{xiong2013supervised, cao2014face, ren2014face, zhu2015face_regression} have been recently proposed to directly estimate landmark locations by applying cascaded regressors to an input image. These methods mostly do not consider the visibility of landmarks under different view angles. Consequently, their performance degrades substantially for non-frontal faces, and their detected landmarks could be ambiguous because the anatomically correct landmarks might be invisible due to self-occlusion (see Fig.~\ref{fig:problem}).

A few methods focused on large-pose face alignment, which can be roughly divided into two categories: multi-view based and 3D model based. Multi-view based methods~\cite{zhu2012face, yu2013pose} define different sets of landmarks as templates, one for each view range. Given an input image, they fit the multi-view templates to it and choose the best fitted one as the final result. These methods are usually complicated to apply, and cannot detect invisible self-occluded landmarks. 3D model based methods, in contrast, can better handle self-occluded landmarks with the assistance of 3D face models. Their basic idea is to fit a 3D face model to the input image to recover the 3D landmark locations. Most of these methods~\cite{jeni2015dense, jourabloo2015pose, zhu2016CVPR, Jourabloo_2016_CVPR, jourabloo2017ijcv,pose-invariant-face-alignment-with-a-single-cnn} use 3D morphable models (3DMM)~\cite{blanz1999morphable} --- either a simplified one with a sparse set of landmarks~\cite{jourabloo2015pose, zhu2016CVPR} or a relatively dense one~\cite{jeni2015dense}. They estimate the 3DMM parameters by using cascaded regressors with texture features as the input. In~\cite{jourabloo2015pose}, the visibility of landmarks is explicitly computed, and the method can cope with face of yaw angles ranging from -$90^{\circ}$ to $90^{\circ}$, whereas the method in~\cite{jeni2015dense} does not work properly for faces of yaw angles beyond $60^{\circ}$. In~\cite{tulyakov2015regressing}, Tulyakov and Sebe propose to directly estimate the 3D landmark locations via texture-feature-based regressors for faces of yaw angles up to $50^{\circ}$. 

These existing 3D model based methods regress between 2D image features and 3D landmark locations (or indirectly, 3DMM parameters). While our proposed approach is also based on 3D model, unlike existing methods, it carries out regressions both on 2D images and in the 3D space. Regressions on 2D images predict 2D landmarks, while regressions in the 3D space predict 3D landmarks coordinates. By integrating both regressions, our proposed method can more accurately estimate landmarks, and better handle self-occluded landmarks. It thus works well for images of arbitrary view angles in [-$90^{\circ}, 90^{\circ}$]. 

\subsection{3D Face Reconstruction}
Estimating the 3D face geometry from a single 2D image is an ill-posed problem. Existing methods, such as Shape from Shading (SFS) and 3DMM, thus heavily depend on priors or constraints. SFS based methods~\cite{kemelmacher20113d, roth2017adaptive} usually utilize an average 3D face model as a reference, and assume the Lambertian lighting model for the 3D face surface. One limitation of SFS methods lies in its assumed connection between 2D texture clues and 3D shape, which could be weak to discriminate among different individuals. 3DMM~\cite{blanz1999morphable, blanz2003face, romdhani2005estimating, hu2017efficient,nonlinear-3d-face-morphable-model} establishes statistical parametric models for both texture and shape, and represents a 3D face as a linear combination of basis shapes and textures. To recover the 3D face from a 2D image, 3DMM-based methods estimate the combination coefficients by minimizing the discrepancy between the input image and the image rendered from the reconstructed 3D face. They can better cope with 2D face images of varying illuminations and poses. However, they still suffer from invisible facial landmarks when the input face has large pose angles. To deal with extreme poses, Lee et~al.~\cite{lee2012single}, Qu et~al.~\cite{qu2014fast} and Liu et~al.~\cite{liu2015cascaded} propose to discard the self-occluded landmarks or treat them as missing data. 

%Tran et al. \cite{tran2016regressing} uses a CNN to regress 3DMM shape and texture parameters for recognition purpose and the method is robust to the facial landmarks.

All the aforementioned 3D face reconstruction methods require landmarks as input. Consequently, they either manually mark the landmarks, or employ standalone face alignment methods to automatically locate the landmarks. Very recently, Tran et~al. \cite{tran2016regressing} propose a convolutional neural network (CNN) based method to estimate discriminative 3DMM parameters directly from single 2D images without requirement of input landmarks.  Yet, existing methods always generate 3D faces that have the same pose and expression as the input image, which may not be desired in face recognition due to the challenge of matching 3D faces with expressions~\cite{drira20133d}. In this paper, we improve 3D face reconstruction by (i) integrating the face alignment step into the 3D face reconstruction procedure, and (ii) reconstructing both expressive and PEN 3D faces, which is shown to be useful for face recognition.

\subsection{Unconstrained Face Recognition}
Face recognition has been developed rapidly in the past decade, especially since the emergence of deep learning techniques. 
Although automated methods~\cite{schroff2015facenet, Sun2015DeepID3, Zhou2015Naive} outperform humans in face recognition accuracy on the labelled faces in the wild (LFW) benchmark database, it is still very challenging to recognize faces in unconstrained images with large poses or intensive expressions~\cite{corneanu2016survey, LuanPose2017}. 
Potential reasons for degraded accuracy on off-angle and expressive faces include (i) off-angle faces usually have less discriminative texture information for identification than frontal ones, resulting in small inter-class differences, (ii) cross-view faces (e.g., frontal and profile faces) may have very limited features in common, leading to large intra-class differences, and (iii) pose and expression variations could cause substantial deformation to faces. 

Existing methods recognize off-angle and expressive faces either by extracting invariant features or by normalizing out the pose or expression deformation. 
Yi et~al.~\cite{yi2013towards} fitted a 3D face mesh to an arbitrary-view face, and extracted pose-invariant features based on the 3D face mesh adaptively deformed to the input face. 
In DeepFace~\cite{taigman2014deepface}, the input face was first aligned to the frontal view with assistance of a generic 3D face model, and then recognized utilizing a deep network. 
Zhu et~al.~\cite{zhu2015high} proposed to generate frontal and neutral face images from the input images by using 3DMM~\cite{blanz1999morphable} and deep convolutional neural networks. 
Very recently, generative adversarial networks (GAN) have been explored by Tran et~al.~\cite{LuanPose2017, representation-learning-by-rotating-your-faces} for unconstrained face recognition. They devised a novel network, namely DR-GAN, which simultaneously synthesizes frontal faces and learn pose-invariant feature representations. Hu et~al.~\cite{hu2017FG} proposed to directly transform a non-frontal face into frontal face by Learning a Displacement Field network (LDF-Net). LDF-Net achieves state-of-the-art performance for face recognition across poses on Multi-PIE, especially at large poses. 
To summarize, all these existing methods carry out pose and expression normalization on 2D faces and utilize merely 2D features for recognition. 
In this paper, on the contrary, we generate pose and expression normalized 3D faces from the input 2D images, and use these resultant 3D faces to improve the unconstrained face recognition accuracy.

\begin{figure}[t]
\begin{center}
\includegraphics[width=3.2in]{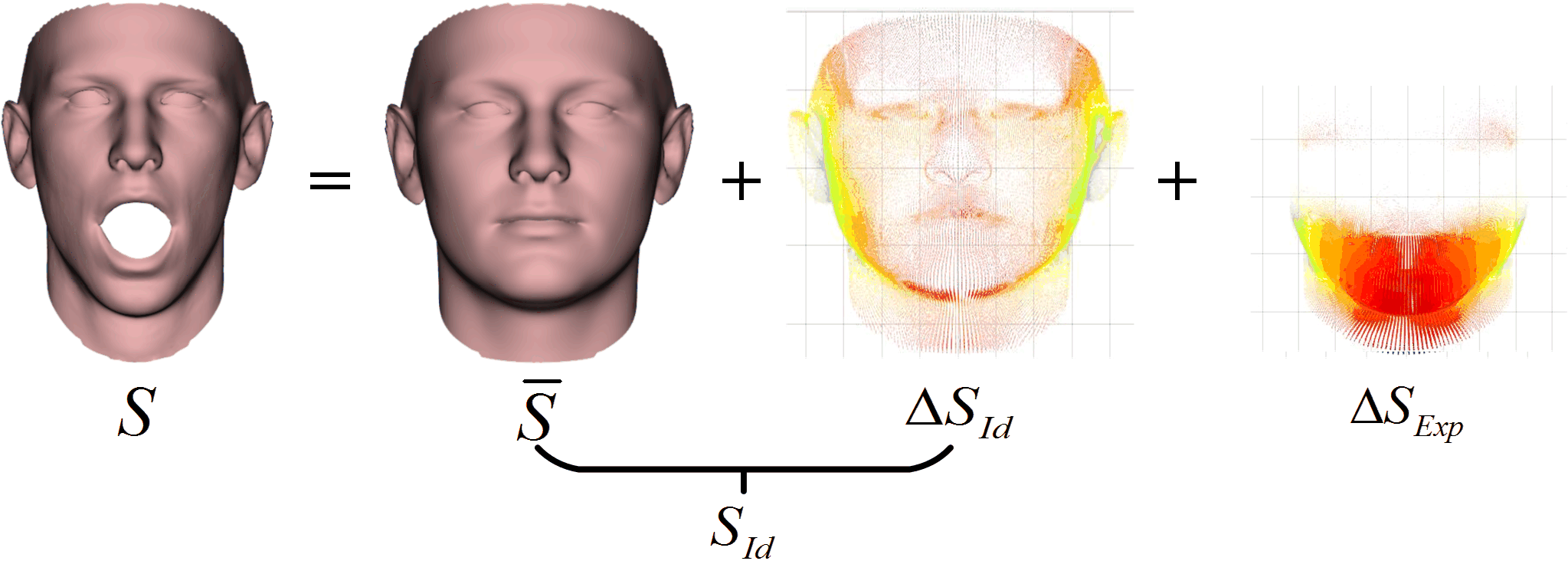}
\end{center}
   \caption{A 3D face shape of a subject ($S$) is represented as summation of the mean pose-and-expression-normalized (PEN) 3D face shape ($\bar{S}$), the difference between the subject's PEN 3D shape and the mean PEN 3D shape ($\Delta S_{Id}$), and the expression deformation ($\Delta S_{Exp}$).}%The defined 3D shape model. Deformations on the face can be categorized into two separate subsets offset: identity and expression parts.
\label{fig:shape_composition}
\end{figure}

\begin{figure*}[t]
\begin{center}
\includegraphics[width=7.0in]{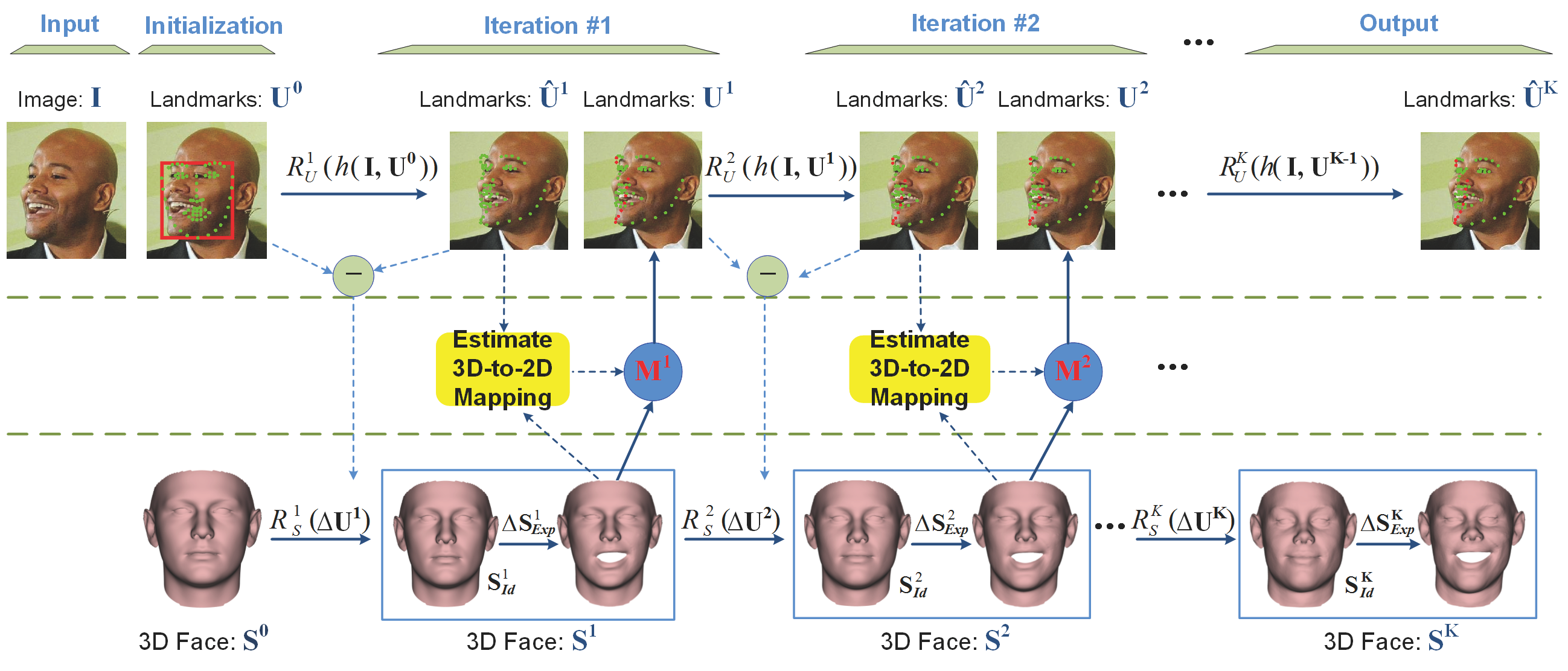}
\end{center}
   \caption{Flowchart of the proposed joint face alignment and 3D face reconstruction method.}
\label{fig:Flowchart}
\end{figure*}

\section{Proposed Method}
\label{sec:method}
In this section, we introduce the proposed joint face alignment and 3D face reconstruction method and its implementations in detail. 
We start by defining the 3D face model with separable identity and expression components, and based on this model formulate the problem of interest. 
We then provide the overall procedure of the proposed method. 
Afterwards, the preparation of training data is presented, followed by the introduction of key steps in the proposed method, including learning 2D landmark and 3D shape regressors, and estimating 3D-to-2D mapping and landmark visibility. 
Finally, a deep learning based nonlinear implementation of the proposed method is given. 

\subsection{Problem Formulation}\label{sec:problem_formulation}
We denote an \emph{n}-vertex frontal pose 3D face shape of one subject as
\begin{equation}
S = \begin{pmatrix}
 x_{1} &\quad x_{2} &\quad \cdots  &\quad x_{n} \\ 
 y_{1} &\quad y_{2} &\quad \cdots  &\quad y_{n}\\ 
 z_{1} &\quad z_{2} &\quad \cdots  &\quad z_{n}
\end{pmatrix} \in \mathbb{R}^{3 \times n},
\end{equation}
and represent it as a summation of three components: 
\begin{equation}
\label{eq:S_model}
S = S_{Id} + \Delta S_{Exp} =\bar{S} + \Delta S_{Id} + \Delta S_{Exp},
\end{equation}
where $\bar{S}$ is the mean of frontal pose and neutral expression 3D face shapes, termed pose-and-expression-normalized (PEN) 3D face shape, $\Delta S_{Id}$ is the difference between the subject's PEN 3D shape (denoted as $S_{Id}$) and $\bar{S}$, and $\Delta S_{Exp}$ is the expression-induced deformation in $S$ w.r.t.~$S_{Id}$ (Fig.~\ref{fig:shape_composition}).

We use $S_{L}$ to denote a subset of $S$ with columns corresponding to $l$ landmarks. 
The projections of these landmarks onto an image $\mathbf{I}$ of the subject with arbitrary view are represented by
\begin{equation}\label{eq:u}
U = \begin{pmatrix}
 u_{1}&\quad u_{2} &\quad \cdots  &\quad u_{l}\\ 
 v_{1}&\quad v_{2} &\quad \cdots  &\quad v_{l}
\end{pmatrix} = f_{C} \circ f_{P} ({S}_{L}) \in \mathbb{R}^{2 \times l},
\end{equation}
where $f_{C}$ and $f_{P}$ are, respectively, camera projection and pose-induced deformation. 
In this work, we employ a 3D-to-2D mapping matrix $\mathbf{M} \approx f_{C}\circ f_{P}$ to approximate the composite effect of pose-induced deformation and camera projection.

Given a face image $\mathbf{I}$, our goal is to simultaneously estimate its landmarks ${U}$,  PEN 3D shape $S_{Id}$, and expression deformation $\Delta S_{Exp}$. 
Note that, in some context, we also write the 3D shape and landmarks as column vectors: $\mathbf{S}=(x_{1}, y_{1}, z_{1}, \cdots, x_{n}, y_{n}, z_{n})^{\mathsf{T}}$, and $\mathbf{U}=(u_{1}, v_{1}, \cdots, u_{l}, v_{l})^{\mathsf{T}}$, where `$\mathsf{T}$' is transpose operator.

\subsection{The Overall Procedure}
Figure~\ref{fig:Flowchart} shows the flowchart of the proposed method. 
Given an image $\mathbf{I}$, its 3D shape $\mathbf{S}$ is initialized as the mean PEN 3D shape of training faces (i.e., $\mathbf{S}^{0}=\bar{\mathbf{S}}$). 
Its landmarks $\mathbf{U}$ are initialized by placing the mean landmarks of training frontal and neutral faces into the face region specified by a bounding box in $\mathbf{I}$ via similarity transforms. 
$\mathbf{U}$ and $\mathbf{S}$ are iteratively updated by applying a series of regressors. 
Each iteration contains three steps: (i) updating landmarks, (ii) updating 3D face shape, and (iii) refining landmarks.

\emph{\textbf{Updating landmarks~~}} This step updates the landmarks' locations from $\mathbf{U}^{k-1}$ to $\hat{\mathbf{U}}^{k}$ based on the texture features in the image. 
This is similar to the conventional cascaded regressor based 2D face alignment~\cite{xiong2013supervised}. 
The adjustment to the landmarks' locations in $k^{\texttt{th}}$ iteration, $\Delta{\mathbf{U}^{k}}$ is determined by the local texture feature around $\mathbf{U}^{k-1}$ via a regressor,
\begin{equation}\label{eq:du}
\Delta{\mathbf{U}}^{k} = R_{U}^{k}(h(\mathbf{I}, \mathbf{U}^{k-1})),
\end{equation}
where $h(\mathbf{I}, \mathbf{U})$ denotes the texture feature extracted around the landmarks $\mathbf{U}$ in the image $\mathbf{I}$, and $R_{U}^{k}$ is a regression function. 
The landmarks can then be updated by $\hat{\mathbf{U}}^{k} = \mathbf{U}^{k-1} + \Delta{\mathbf{U}}^{k}$. 
The method for learning these landmark regressors in linear case will be introduced in Sec.~\ref{sec:learnlandmark}.

\emph{\textbf{Updating 3D face shape~~}} In this step, the aforementioned landmark location adjustment is used to estimate the  adjustment of the 3D shape $\Delta{\mathbf{S}}^{k}$, which consists of two components, $\Delta{\mathbf{S}}_{Id}^{k}$ and $\Delta{\mathbf{S}}_{Exp}^{k}$. Specifically, a regression function $R_{S}^{k}$ models the correlation between the landmark location adjustment $\Delta{\mathbf{U}}^{k}$ and the expected adjustment $\Delta{\mathbf{S}}_{Id}^{k}$ and $\Delta{\mathbf{S}}_{Exp}^{k}$,  i.e.,
\begin{equation}\label{eq:ds}
\Delta{\mathbf{S}}^{k} = [\Delta{\mathbf{S}}_{Id}^{k}; \Delta{\mathbf{S}}_{Exp}^{k}] = R_{S}^{k}(\Delta{\mathbf{U}}^{k}).
\end{equation}
The 3D shape can be then updated by $\mathbf{S}^{k} = \mathbf{S}^{k-1} + \Delta{\mathbf{S}}_{Id}^{k} + \Delta{\mathbf{S}}_{Exp}^{k}$. The method for learning these shape regressors in linear case will be given in Sec.~\ref{sec:learnshape}.

\emph{\textbf{Refining landmarks~~}} Once a new estimate of the 3D shape is obtained, the landmarks can be further refined with the assitance of the 3D-to-2D mapping matrix. 
We estimate $\mathbf{M}^{k}$ based on $\mathbf{S}^{k}$ and $\hat{\mathbf{U}}^{k}$. 
The refined landmarks $\mathbf{U}^{k}$ can be obtained by projecting $\mathbf{S}^{k}$ onto the image via $\mathbf{M}^{k}$ according to Eq.~(\ref{eq:u}). 
In this process, the landmark visibility is also re-computed. Details of this step will be given in Sec.~\ref{sec:learnmap}.

\subsection{Training Data Preparation} \label{sec:train_data_preparation}
Before we provide details of the three steps, we first introduce the training data needed for learning the landmark and shape regressors, which will also facilitate the understanding of our algorithms. 
Since the purpose of these regressors is to gradually adjust the estimated landmark and shape towards their ground truth, we need a sufficient number of triplet data $\{(\mathbf{I}_{i}, \mathbf{S}^{*}_{i}, \mathbf{U}^{*}_{i})_{i=1}^N\}$, where $\mathbf{S}^{*}_{i}$ and $\mathbf{U}^{*}_{i}$ are, respectively, the ground truth 3D shape and landmarks for the image $\mathbf{I}_{i}$, and $N$ is the total number of training samples. 
All the 3D shapes have established dense correspondences among their vertices; i.e., they have the same number of vertices, and vertices of the same index in the 3D shapes have the same semantic meaning. 
Here, each of the ground truth 3D shapes includes two parts, the PEN 3D shape $\mathbf{S}^{*}_{Id}$ and its expression shape $\mathbf{S}^{*}_{Exp} = \bar{\mathbf{S}} + \Delta \mathbf{S}^{*}_{Exp}$, i.e., $\mathbf{S}^{*} = [\mathbf{S}^{*}_{Id}; \mathbf{S}^{*}_{Exp}]$. Moreover, both visible and invisible landmarks in $\mathbf{I}_{i}$ have been annotated and included in $\mathbf{U}^{*}_{i}$. For invisible landmarks, the annotated positions should be anatomically correct positions (e.g., the red points in Fig.~\ref{fig:problem}).

Obviously, to enable regressors to cope with expression and pose variations, the training data should contain faces of these variations. 
It is, however, difficult to find in the public domain such data sets of 3D faces and corresponding annotated 2D images with various expressions/poses. 
Thus, we construct two  training sets by ourselves: one based on BU3DFE~\cite{yin20063d}, and the other based on 300W-LP~\cite{Sagonas2013300, zhu2016CVPR}.

\textbf{BU3DFE} database contains 3D face scans of $56$ females and $44$ males, acquired in neutral plus six basic expressions (happiness, disgust, fear, anger, surprise and sadness). 
All basic expressions are acquired at four intensity levels. 
These 3D scans have been manually annotated with $84$ landmarks ($83$ landmarks provided by the database plus one nose tip marked by ourselves). 
For each of the $100$ subjects, we select the scans of neutral and the level-one intensity of the rest six expressions as the ground truth 3D face shapes. 
From each of the chosen seven scans of a subject, $19$ face images are rendered at different poses (-$90^{\circ}$ to $90^{\circ}$ yaw with a $10^{\circ}$ interval) with landmark locations recorded. 
As a result, each subject has $133$ images of different poses and expressions. 
We use the method in~\cite{bolkart20153d} to establish dense correspondence of the 3D scans of $5{,}996$ vertices. 
With the registered 3D scans, we compute the mean PEN 3D face shape by averaging all the subjects' PEN 3D shapes, which are defined by their 3D scans of frontal pose and neutral expression. 
All the images of one subject share the same PEN 3D shape of that subject, while their expression shapes can be obtained by first subtracting from their corresponding 3D scans, their PEN 3D face shape, and then adding the mean PEN 3D shape.

\begin{figure}[t]
\centering
\includegraphics[width=\linewidth]{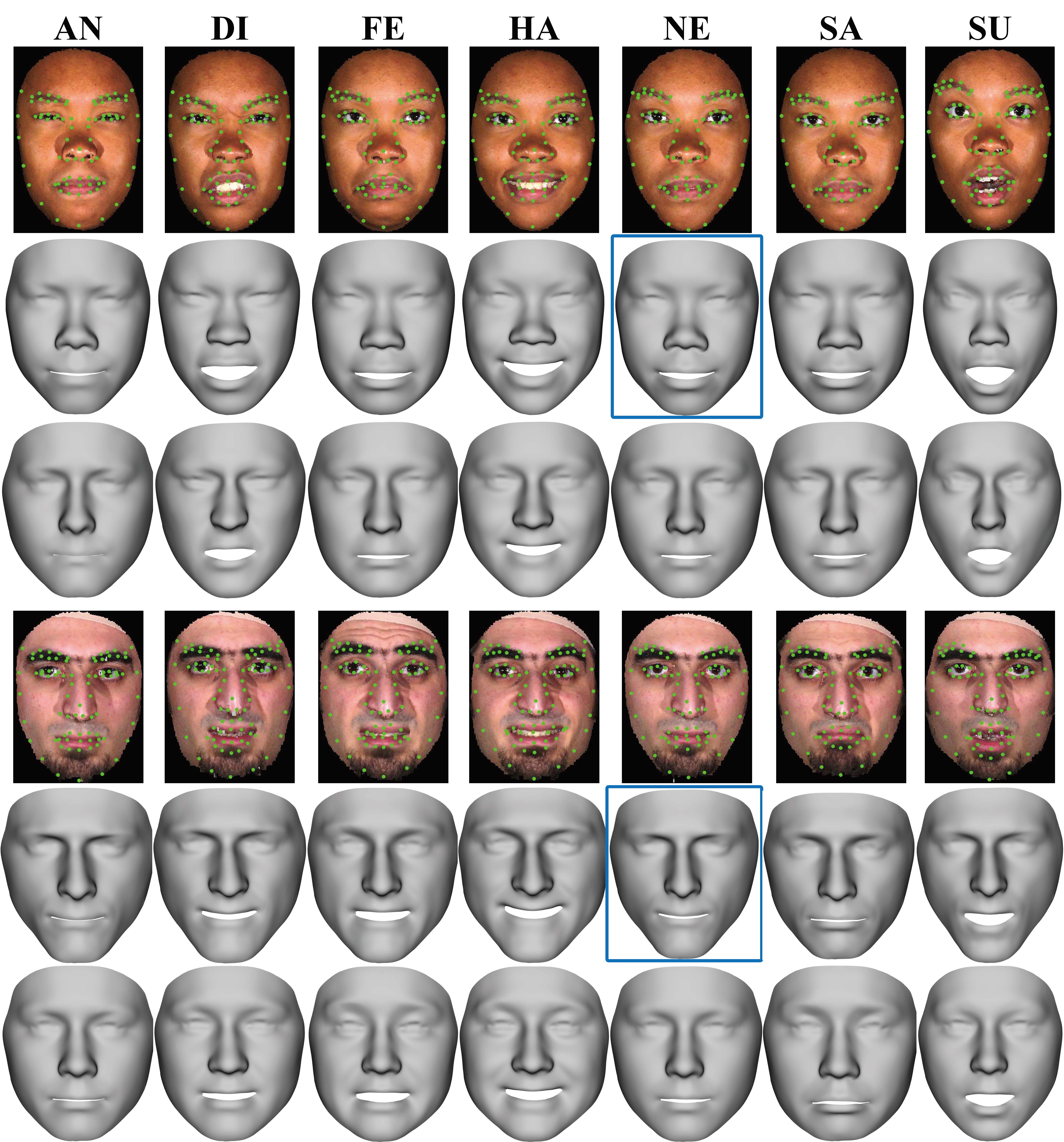}
\caption{Example images with annotated landmarks ($1^{st}$, $4^{th}$ rows), their 3D faces ($2^{nd}$, $5^{th}$ rows) and expression shapes ($3^{rd}$, $6^{th}$ rows) from the BU3DFE database. Seven expressions are shown: angry (AN), disgust (DI), fear (FE), happy (HA), neutral (NE), sad (SA), and surprise (SU). The 3D shapes corresponding to the neutral expression are their PEN 3D face shapes, which are highlighted in blue boxes.} %Example 2D face images and corresponding registered 3D shapes with different expressions from the BU3DFE database.
\label{fig:aligned_bu3d}
\end{figure}

\begin{figure}[t]
\centering
\includegraphics[width=3.55in]{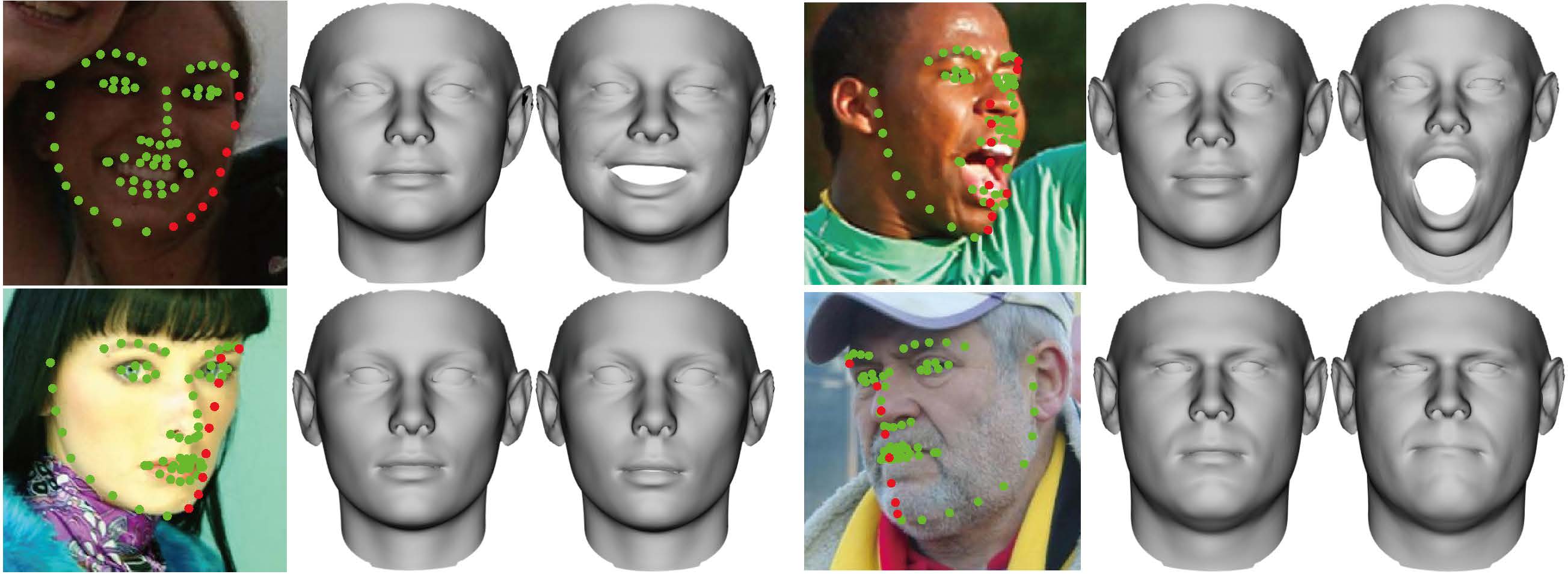}
\caption{Four subjects in 300W-LP. From left to right: images with annotated landmarks, PEN 3D face shapes, and expression shapes.}%Example 2D face images and corresponding 3D face shapes from the 300W-LP database.
\label{fig:300wp_example}
\end{figure}

\textbf{300W-LP} database~\cite{zhu2016CVPR} is created based on 300W~\cite{Sagonas2013300} database, which integrates multiple face alignment benchmark datasets (i.e., AFW~\cite{zhu2012face}, LFPW~\cite{Belhumeur2011Localizing}, HELEN~\cite{Zhou2013Extensive}, IBUG~\cite{Sagonas2013300} and XM2VTS~\cite{messer1999xm2vtsdb}). 
It includes $122{,}450$ in-the-wild images of a wide variety of poses and expressions. 
For each image, its corresponding registered PEN 3D shape and expression shape are estimated by using the method in~\cite{zhu2015high} based on BFM~\cite{paysan20093d} and FaceWarehouse~\cite{cao2014facewarehouse}. 
The obtained 3D faces have $53{,}215$ vertices. %FIXME $n=5,996$! use a different variable. Otherwise n has two values.
Figure~\ref{fig:aligned_bu3d} and~\ref{fig:300wp_example}  shows example images and corresponding PEN 3D shapes and expression shapes in our training sets.

\subsection{Learning Landmark Regressors}\label{sec:learnlandmark}
According to Eq.~(\ref{eq:du}), landmark regressors estimate the adjustment to $\mathbf{U}^{k-1}$ such that the updated landmarks $\mathbf{U}^{k}$ are closer to their ground truth, which, along with landmark visibility, are given by $\mathbf{U}^{*}$ in training.
%In the training phase, the true positions and visibility of the landmarks are given by the ground truth $\mathbf{U}^{*}$. 
Therefore, the objective of landmark regressors $R^{k}_{U}$ is to better predict the difference between $\mathbf{U}^{k-1}$ and $\mathbf{U}^{*}$. 
In this section, we first implement the proposed method in a linear manner, by optimizing:
\begin{equation}\label{eq:2d_object}
R^{k}_{U} = \mathop{\arg \min}  \limits_{R^{k}_{U}}\sum_{i=1}^N\parallel \left(\mathbf{U}^{*}_{i}  - \mathbf{U}^{k-1}_{i}\right) - R^{k}_{U}(h(\mathbf{I}_{i}, \mathbf{U}_{i}^{k-1}))\parallel_2^2,
\end{equation}
which has a closed-form least-square solution. 
Note that, as we will show later, other nonlinear regression schemes, such as CNN~\cite{Jourabloo_2016_CVPR}, can also be adopted in our framework.  

We use $128$-dim SIFT descriptors~\cite{lowe2004distinctive} as the local feature. 
The feature vector of $h$ is a concatenation of the SIFT descriptors at all the $l$ landmarks, i.e., a $128l$-dim vector. 
If a landmark is invisible, no feature will be extracted, and its corresponding entries of $h$ will be zero. 
Note that the regressors estimate the semantic locations of all landmarks including invisible ones. 

\subsection{Learning 3D Shape Regressors}\label{sec:learnshape}
The landmark adjustment $\Delta{\mathbf{U}}^{k}$ is also used as the input to the 3D shape regressor $R^{k}_{S}$. 
The objective of $R^{k}_{S}$ is to compute an update to the initially estimated 3D shape $\mathbf{S}^{k-1}$ in the $k^{\texttt{th}}$ iteration to minimize the difference between the updated 3D shape and the ground truth. 
Using similar linear regressors, the 3D shape regressors can be learned by solving the following optimization via least squares:
\begin{equation}\label{eq:3d_object}
R^{k}_{S} = \mathop {\arg \min } \limits_{{R^{k}_S}}\sum_{i=1}^N\parallel (\mathbf{S}^{*}_{i} - \mathbf{S}^{k-1}_{i})
- R^{k}_{S}\left(\Delta{\mathbf{U}}^{k}_{i}\right)\parallel_2^2,
\end{equation}
with its closed-form solution as
\begin{equation}
\label{eqn::answer2}
 R^{k}_{S}=\Delta \mathbb{S}^{k}(\Delta \mathbb{U}^{k})^{\mathsf{T}}(\Delta \mathbb{U}^{k}(\Delta \mathbb{U}^{k})^{\mathsf{T}})^{-1},
 \end{equation}
where $\Delta \mathbf{\mathbb{S}}^{k} = \mathbf{\mathbb{S}}^{*}-\mathbf{\mathbb{S}}^{k-1}$ and $\Delta \mathbf{\mathbb{U}}^{k}$ are, respectively, the 3D shape and landmark adjustment. $\mathbb{S}$ and $\mathbb{U}$ denote, respectively, the ensemble of 3D face shapes and 2D landmarks of all training samples with one column per sample.

Since $\mathbb{S} \in \mathbb{R}^{6n\times N}$ (recall that $\mathbb{S}$ has two parts, PEN shape and expression deformation) and $\mathbb{U}\in \mathbb{R}^{2l\times N}$, it can be mathematically shown that $N$ should be larger than $2l$ so that $\Delta \mathbb{U}^{k}(\Delta \mathbb{U}^{k})^{\mathsf{T}}$ is invertible. 
Fortunately, since the  landmark set is usually sparse, this requirement can be easily satisfied in real-world applications.

\subsection{3D-to-2D Mapping and Landmark Visibility}\label{sec:learnmap}
In order to refine landmarks with the updated 3D shape, we project the 3D shape to the 2D image with a 3D-to-2D mapping matrix. 
In this paper, we dynamically estimate the mapping matrix based on $\mathbf{S}^{k}$ and $\hat{\mathbf{U}}^{k}$. 
As discussed in Sec.~\ref{sec:problem_formulation}, the mapping matrix is a composite effect of pose-induced deformation and camera projection. 
By assuming a weak perspective camera projection as in prior work~\cite{Zhou_2015_CVPR, jourabloo2015pose}, the mapping matrix $\mathbf{M}^{k}$ is represented by a $2\times 4$ matrix, and can be estimated as a least-square solution to the following fitting problem:
\begin{equation}\label{eq:compute_m}
\mathbf{M}^{k} = \mathop {\arg \min } \limits_{\mathbf{M}^{k}} \parallel \hat{{U}}^{k} - \mathbf{M}^{k} {S}^{k}_{L}\parallel_2^2.
\end{equation}
Once a new mapping matrix is computed, the landmarks can be further refined as ${U}^{k} = \mathbf{M}^{k}{S}^{k}_{L}$.

The visibility of the landmarks can be then computed based on the mapping matrix $\mathbf{M}$ using the method in~\cite{jourabloo2015pose}. Suppose the average surface normal around a landmark in the 3D face shape $\mathbf{S}$ is $\overrightarrow{\textbf{n}}$. 
Its visibility $\textbf{v}$ is measured by 
\begin{equation}\label{eq::visibility}
\textbf{v} = \frac{1}{2}\left(1 + \rm{sgn}\left(\overrightarrow{\textbf{n}} \cdot \left( \frac{\mathbf{M}_{1}}{\left \| \mathbf{M}_{1} \right \|}\times \frac{\mathbf{M}_{2}}{\left \| \mathbf{M}_{2} \right \|} \right)\right)\right),
\end{equation}
where $\rm{sgn}()$ is the sign function, `$\cdot$' means dot product and `$\times$' cross product, and $\mathbf{M}_{1}$ and $\mathbf{M}_{2}$ are the left-most three elements at the top two rows of $\mathbf{M}$. 
This rotates the surface normal and validates if it points toward the camera.

Algorithm~\ref{alg:reconstruction_algorithm} summarizes the process of learning the cascaded coupled linear regressors. 
Next, we introduce an alternative implementation of our proposed method by using nonlinear regressors, i.e., neural networks.

%The whole process of learning the cascaded coupled landmark and 3D shape regressors is summarized in Algorithm \ref{alg:reconstruction_algorithm}.

\begin{algorithm}[t]
\caption{Learning Cascaded Coupled Linear Regressors}
\label{alg:reconstruction_algorithm}
\begin{algorithmic}[1]
\REQUIRE Training data $\{(\mathbf{I}_{i}, \mathbf{S}^{*}_{i}, \mathbf{U}^{*}_{i})\vert i=1,2,\cdots,N\}$, initial shape $\mathbf{S}^{0}_{i}$ \& landmarks $\mathbf{U}^{0}_{i}$.
\ENSURE Cascaded coupled-regressors $\left \{R^{k}_{U},R^{k}_{S}\right \}_{k=1}^{K}$.
%\STATE Compute initial 3D-to-2D mapping matrix $\mathbf{M}^{0}_{i}$ via Eq. (\ref{eq:compute_m}) for all images;
\FOR{$k=1,...,K$} 
\STATE Estimate $R_{U}^{k}$ via Eq.~(\ref{eq:2d_object}), and compute landmark adjustment $\Delta \mathbf{U}^{k}_{i}$ via Eq.~(\ref{eq:du});
\STATE Update landmarks $\hat{\mathbf{U}}^{k}_{i}$ for all images: $\hat{\mathbf{U}}^{k}_{i}=\mathbf{U}^{k-1}_{i}+\Delta \mathbf{U}^{k}_{i}$;
\STATE Estimate $R_{S}^{k}$ via Eq.~(\ref{eq:3d_object}), and compute shape adjustment $\Delta \mathbf{S}^{k}_{i}$ via Eq.~(\ref{eq:ds});
\STATE Update 3D face $\mathbf{S}_{i}^{k}$: $\mathbf{S}_{i}^{k}=\mathbf{S}_{i}^{k-1}+\Delta \mathbf{S}^{k}_{i}$;
\STATE Estimate the 3D-to-2D mapping matrix $\mathbf{M}_{i}^{k}$ via Eq.~(\ref{eq:compute_m});
\STATE Compute the refined landmarks $\mathbf{U}^{k}_{i}$ via Eq.~(\ref{eq:u}) and their visibility via Eq.~(\ref{eq::visibility}).
\ENDFOR 
\end{algorithmic}
\end{algorithm}

\begin{figure}[t]
\centering
\includegraphics[width=\linewidth]{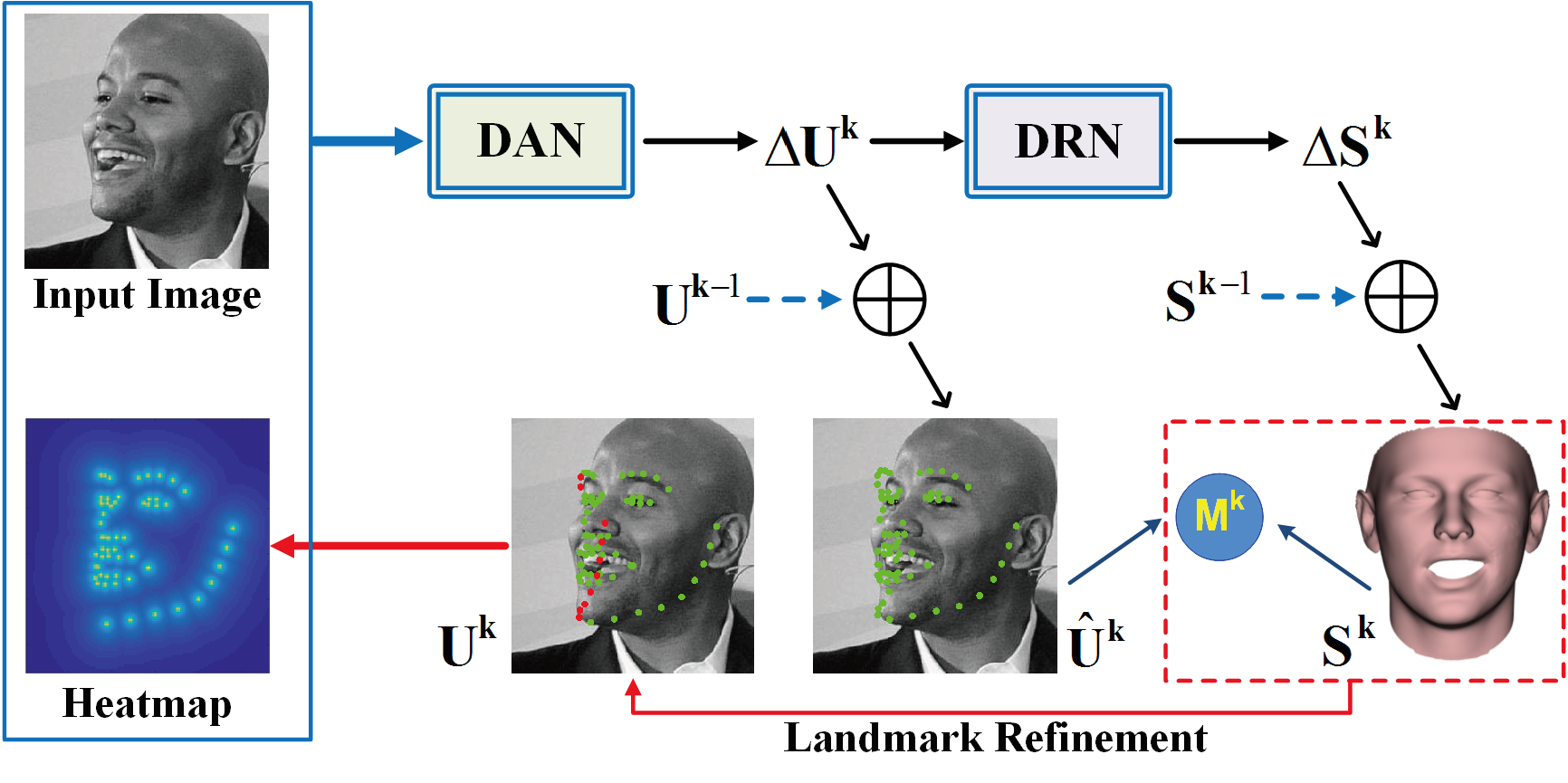}
\caption{Diagram of the proposed method implemented with nonlinear regressors. Deep Alignment Network (DAN) denotes the DCNN-based landmark regressors and Deep Reconstruction Network (DRN) denotes the MLP-based 3D shape regressors. Note that the landmark heatmap is not used at the initial stage.}
\label{fig:non_linear_network}
\end{figure}

\subsection{Nonlinear Regressors}

In the above linear implementation, linear regressors with hand-crafted features are used. Here, we provide a nonlinear implementation, in which landmark and 3D shape regressors are implemented by deep convolutional neural networks (DCNN) and multiple layer perceptions (MLP), respectively. 
Figure~\ref{fig:non_linear_network} shows its pipeline.

Given a face image, as in linear implementation, its landmarks and 3D shape are initialized as the average landmarks and the average 3D shape. 
In every iteration, a landmark heatmap $\mathbf{H}$, which has the same dimension as the input image, is generated from the current estimated landmarks. 
The value of pixel $(p_{u}, p_{v})$ in the heatmap is set as the accumulated contributions of the visible landmarks, and the contribution of a landmark $U_{j}$ is determined by
\begin{equation}
\mathbf{H}(p_{u}, p_{v}) = 1/(1+\min_{U_{j}\in U}\left \| (p_{u}, p_{v})-U_{j} \right \|).
\end{equation}
The heatmap and face image are stacked together as input to the DCNN-based landmark regressor. 
In this paper, we employ the structure of Deep Alignment Network (DAN)~\cite{kowalski2017deep}, and adapt its output layer so that landmark adjustment is estimated. 
The obtained landmark adjustment is then fed into the MLP-based 3D shape regressor (Deep Reconstruction Network, or DRN). 
DRN, consisting of a full-connection layer and a tanh() activation function, computes the 3D shape adjustment. 
After updating the 3D shape with the shape adjustment, we further refine the landmarks as in Sec.~$3.6$.

The DCNN- and MLP-based regressors are learned iteratively. 
We first train the regressors in prior iteration until convergence, and then move on to the next iteration. 
We employ the Euclidean loss in training both regressors.

\section{Application to Face Recognition}\label{sec:facerecognition}

\begin{figure}[t]
\centering
\includegraphics[width=\linewidth]{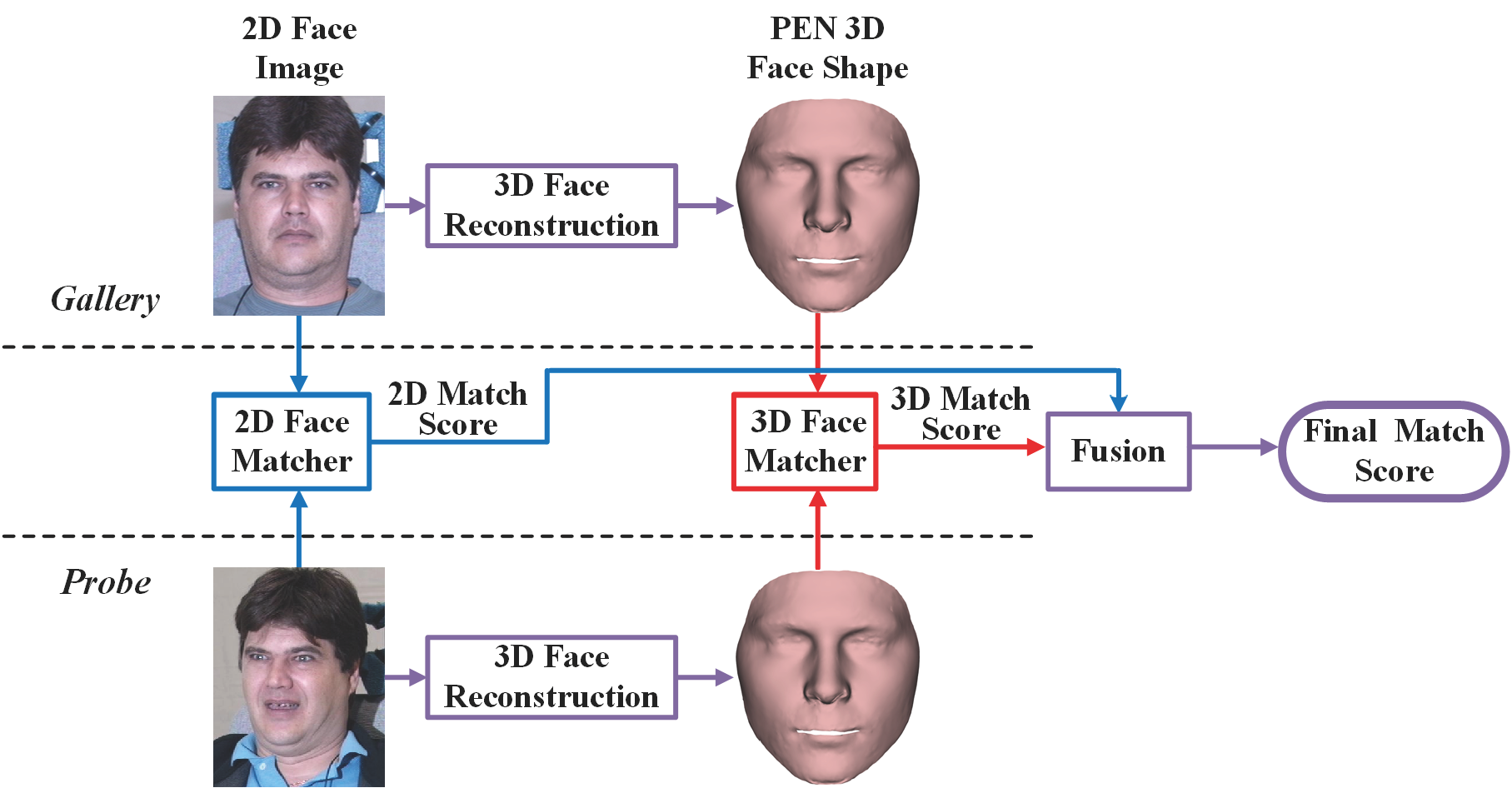}
\caption{Block diagram of the proposed 3D-enhanced face recognition.}
\label{fig:recognition_protocols}
\end{figure}

\begin{figure*}[!t]
\centering
\includegraphics[width=0.98\linewidth]{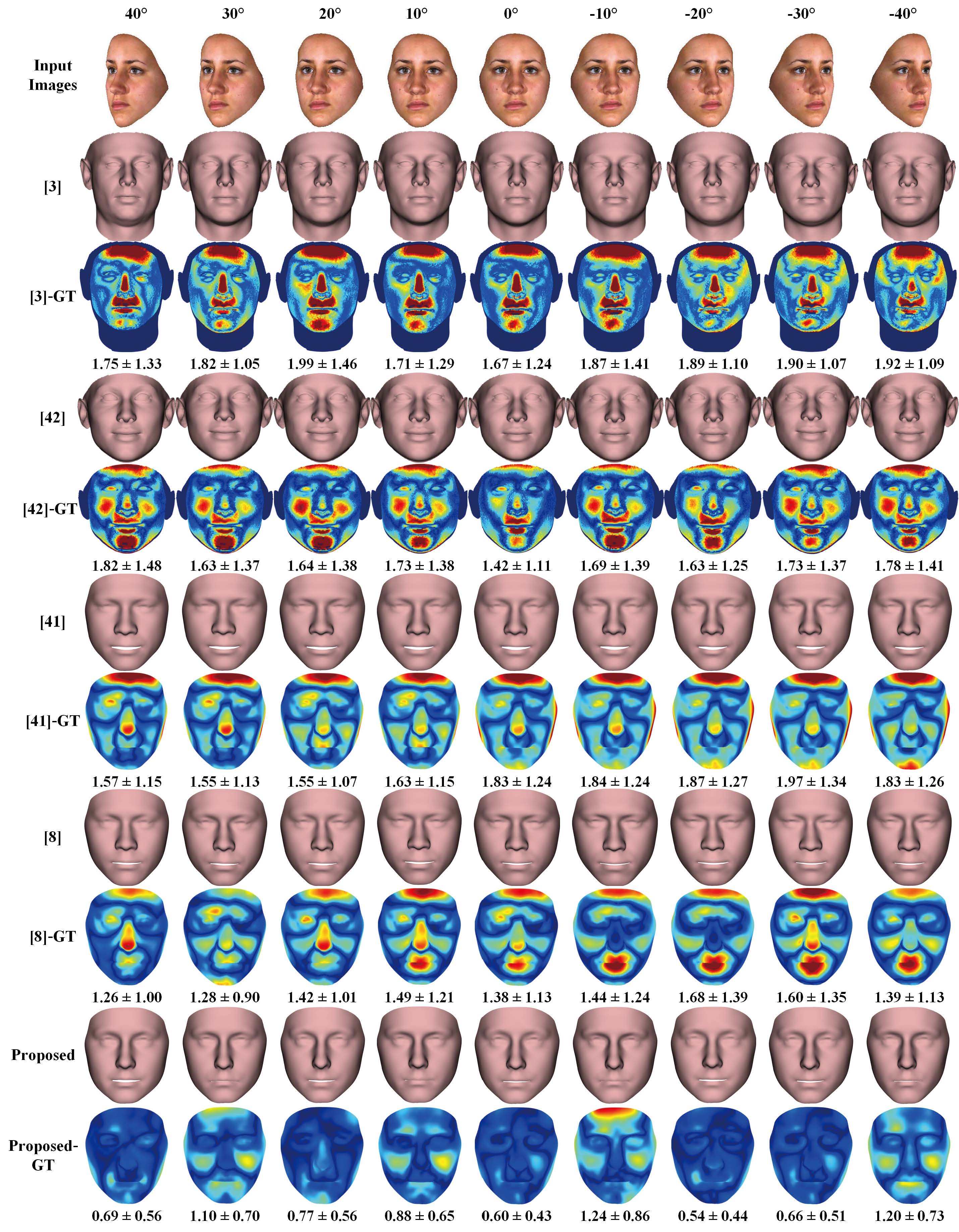}
%\caption{Reconstruction results for a BU3DFE subject at nine different poses. First row: The input images. Second, forth, sixth, eighth and tenth rows: The reconstructed 3D face shapes by \cite{zhu2015high}, \cite{tran2016regressing}, \cite{liu2015cascaded}, \cite{liu2016joint} and the proposed method. Third, fifth, seventh, ninth and eleventh rows: Their corresponding NPDE maps. The colormap goes from dark blue to dark red (corresponding to errors between $0$ and $5$). The numbers under each error map represent the mean and standard deviation (in $\%$).}
\caption{Reconstruction results for a BU3DFE subject at nine poses. The even rows show the reconstructed 3D faces by \cite{zhu2015high}, \cite{tran2016regressing}, \cite{liu2015cascaded}, \cite{liu2016joint} and the proposed method. Except the first row, the odd rows show their corresponding NPDE maps. The colormap goes from dark blue to dark red (corresponding to errors between $0$ and $5$). The numbers under each error map represent the mean and standard deviation (in $\%$).}
\label{fig:bu3d_pose_example}
\end{figure*}

\begin{table*}[t]
\renewcommand{\arraystretch}{0.9}
\centering 
\caption{3D face reconstruction accuracy (MAE) of the proposed method and state-of-the-art methods at different yaw poses on the BU3DFE database.} 
\begin{tabular*}{15.0cm}{l c c c c c c c c c c c}
\toprule
%& \multicolumn{9}{c}{\textbf{Yaw Rotation}} \\ % Amalgamating several columns into one cell is done using the \multicolumn command as seen on this line
%\cmidrule(l){2-12} % Horizontal line spanning less than the full width of the table - you can add (r) or (l) just before the opening curly bracket to shorten the rule on the left or right side
\textbf{Method} & $\pm90^{\circ}$ & $\pm80^{\circ}$ & $\pm70^{\circ}$ & $\pm60^{\circ}$ & $\pm50^{\circ}$ & $\pm40^{\circ}$ & $\pm30^{\circ}$ & $\pm20^{\circ}$ & $\pm10^{\circ}$ & $0^{\circ}$ &\textbf{Avg.}\\ % Column names row
\midrule % In-table horizontal line
\midrule % In-table horizontal line
 Zhu et~al. \cite{zhu2015high} & - & - & - & - & - & $2.73$ & $2.74$ & $2.56$ & $2.32$ & $2.22$ & $2.51$\\ 
 Tran et~al. \cite{tran2016regressing} & - & - & - & - & - & $2.26$ & $2.19$ & $2.16$ & $2.08$ & $2.06$ & $2.15$\\ 
 Liu et~al. \cite{liu2015cascaded} & $1.95$ & $1.91$ & $1.95$ & $1.96$ & $1.97$ & $1.97$ & $1.96$ & $1.98$ & $2.01$ & $2.03$ & $1.97$ \\ 
 Liu et~al. \cite{liu2016joint} & $1.92$ &  $1.89$ & $1.90$ & $1.93$ & $1.95$ & $1.93$ & $1.93$ & $1.95$ &  $1.98$ & $2.01$ & $1.94$\\ 
\midrule % In-table horizontal line
Proposed (Linear) &  $\bf{1.85}$ & $\bf{1.83}$ &  $\bf{1.83}$ &  $\bf{1.83}$ &  $\bf{1.86}$ &  $\bf{1.89}$ &  $\bf{1.90}$ &  $\bf{1.91}$  & $\bf{1.90}$  &  $\bf{1.91}$ & $\bf{1.87}$ \\ 
Proposed (Nonlinear) &  $1.92$ & $1.91$ &  $1.93$ &  $1.92$ &  $1.92$ &  $1.91$ &  $1.92$ &  $1.92$  & $1.93$  &  $1.93$ & $1.92$ \\ 
\bottomrule % Bottom horizontal line
\end{tabular*}
\label{tab:bu3d_pose_compare}
\end{table*}

In this section we apply the reconstructed 3D faces to improve face recognition accuracy on off-angle and expressive faces. 
The basic idea is to utilize the additional feature provided by the reconstructed PEN 3D faces and fuse it with conventional 2D face matchers. Figure~\ref{fig:recognition_protocols} shows the proposed 3D-enhanced face recognition method. 
As can be seen, 3D face reconstruction methods are applied to both gallery and probe faces to generate PEN 3D faces. 
The iterative closest point (ICP) algorithm~\cite{chen1991object} is applied to match the reconstructed normalized 3D face shapes. 
It aligns the 3D shapes reconstructed from probe and gallery images, and computes their distances, which are then converted to similarity scores via subtracting them from the maximum distance. 
These scores are finally normalized to the range of $[0, 1]$ via min-max normalization, and fused with the scores of the conventional 2D face matcher (which are within $[0, 1]$ also) by a sum rule. 
The recognition result for a probe is defined as the subject whose gallery sample has the highest match score with it. 
Note that we employ the ICP-based 3D face matcher and the sum fusion rule for simplicity. 
Other more elaborated 3D face matchers and fusion rules can also be  applied with our proposed method. 
Thanks to the additional discriminative feature in PEN 3D face shapes and its robustness to pose and expression variations, the accuracy of conventional 2D face matchers on off-angle and expressive face images can be effectively improved after fusion with the PEN 3D face based matcher. 
In the next Section, we will experimentally demonstrate this.

\section{Experiments}
\label{sec:exp}
We conduct three sets of experiments to evaluate the proposed method in 3D face reconstruction, face alignment, and face recognition.

\subsection{3D Face Reconstruction Accuracy}
To evaluate the 3D shape reconstruction accuracy, a $10$-fold cross validation is applied to split the BU3DFE data into training and testing subsets, resulting in $11{,}970$ training and $1{,}330$ testing samples. 
We compare the proposed method with its preliminary version in~\cite{liu2016joint} and three state-of-the-art methods in~\cite{liu2015cascaded,zhu2015high,tran2016regressing}. 
The methods in~\cite{liu2016joint,tran2016regressing} reconstruct PEN 3D faces only, while the methods in~\cite{liu2015cascaded,zhu2015high} reconstruct 3D faces that have the same pose and expression as the input images. 
Moreover, the method in~\cite{liu2015cascaded} requires that visible landmarks are available together with the input images. 
In the following experiments, we use the visible landmarks projected from ground truth 3D faces for \cite{liu2015cascaded}. 
For the methods of~\cite{zhu2015high,tran2016regressing}, we use the implementation provided by the authors. 
In the implementation, these two methods are based on the $68$ landmarks that are detected by using~\cite{kazemi2014one}. 
As a result, they cannot be applied to faces of large poses (i.e., beyond $40$ degrees).

We use two metrics to evaluate the 3D face reconstruction accuracy: Mean Absolute Error (MAE) and Normalized Per-vertex Depth Error (NPDE). MAE is defined as~\cite{lei2008face}:
\begin{equation}
\texttt{MAE} = \frac{1}{N_{T}}\sum_{i=1}^{N_{T}}(\| \mathbf{S}^{*}_{i}-\hat{\mathbf{S}}_{i} \|/n), 
\end{equation}
where $N_{T}$ is the total number of testing samples, $\mathbf{S}^{*}_{i}$ and $\hat{\mathbf{S}}_{i}$ are the ground truth and reconstructed 3D face shape of the $i^{\texttt{th}}$ testing sample. 

NPDE measures the depth error at the $j^{\texttt{th}}$ vertex in a testing sample as~\cite{kemelmacher20113d}:
\begin{equation}
\texttt{NPDE}(x_{j}, y_{j}) = \left(|z^{*}_{j} - \hat{z}_{j}|\right)/\left(z^{*}_{max} - z^{*}_{min}\right), 
\end{equation}
where $z^{*}_{max}$ and $z^{*}_{min}$ are the maximum and minimum depth values in the ground truth 3D face of testing samples, and $z^{*}_{j}$ and $\hat{z}_{j}$ are the ground truth and reconstructed depth values at the $j^{\texttt{th}}$ vertex. 
We first report the results of our linear implementation, and then those of the nonlinear one. 
Note that when we mention the proposed method, the linear implementation is referred unless specified.

\begin{figure}[t]
\centering
\includegraphics[width=2.7in]{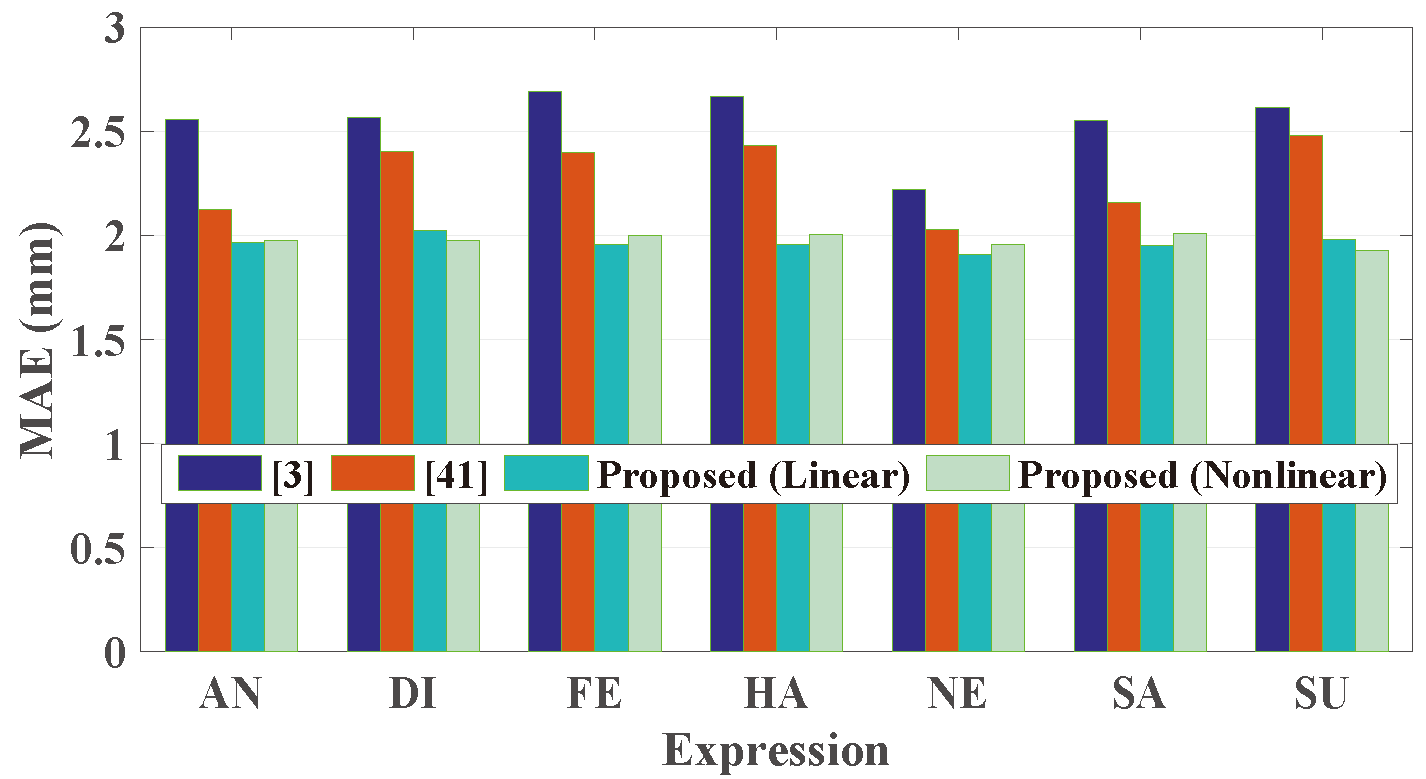}
\caption{3D face reconstruction accuracy (MAE) of the proposed method, \cite{liu2015cascaded} and \cite{zhu2015high} under different expressions: angry (AN), disgust (DI), fear (FE), happy (HA), neutral (NE), sad (SA) and surprise (SU).}
\label{fig:bu3d_exp_error_arxiv}
\end{figure}

\begin{figure}[t]
\centering
\includegraphics[width=2.7in]{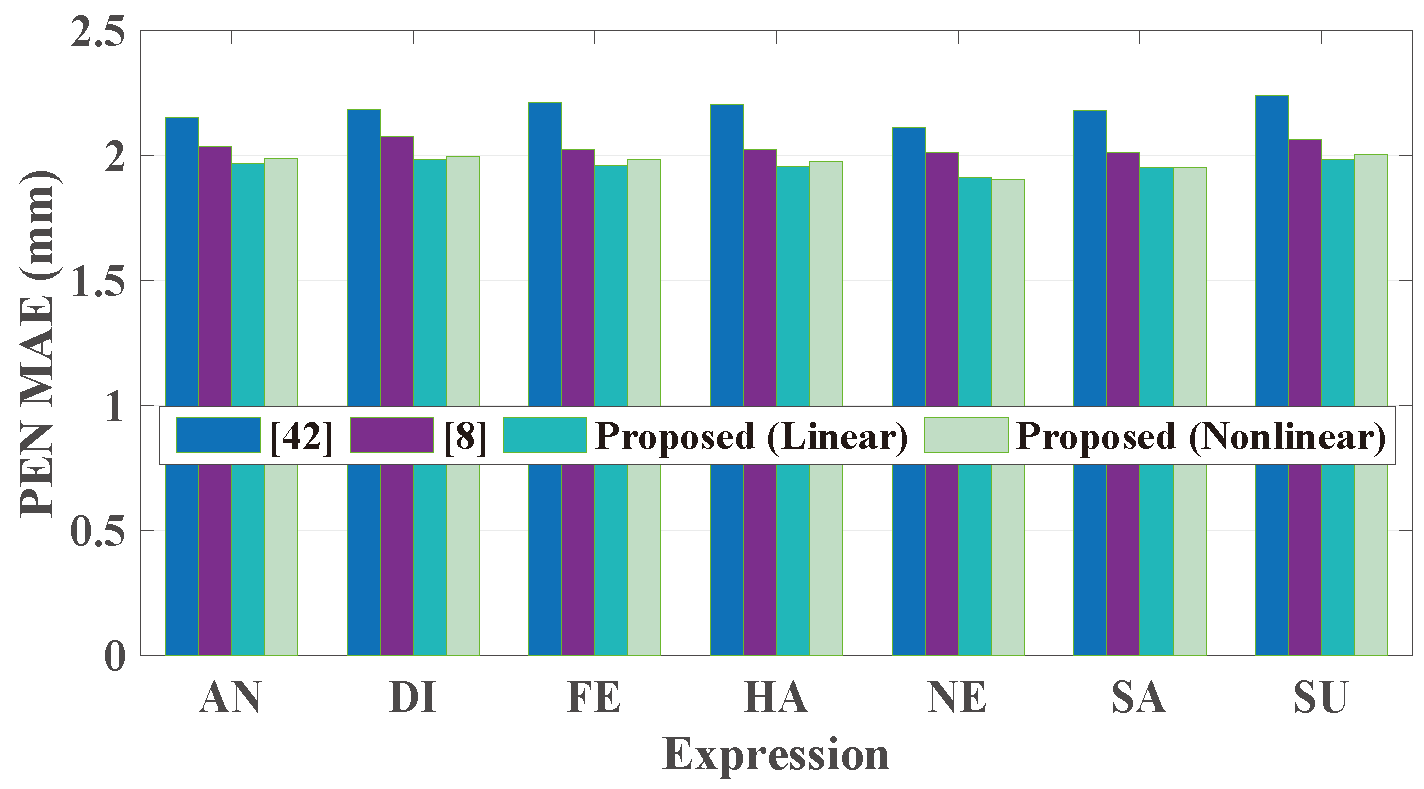}
\caption{PEN 3D face reconstruction accuracy (MAE) of the proposed method, \cite{liu2016joint} and \cite{tran2016regressing} under different expressions.}
\label{fig:bu3d_exp_error_eccv}
\end{figure}

\begin{figure*}[!t]
\centering
\includegraphics[width=5in]{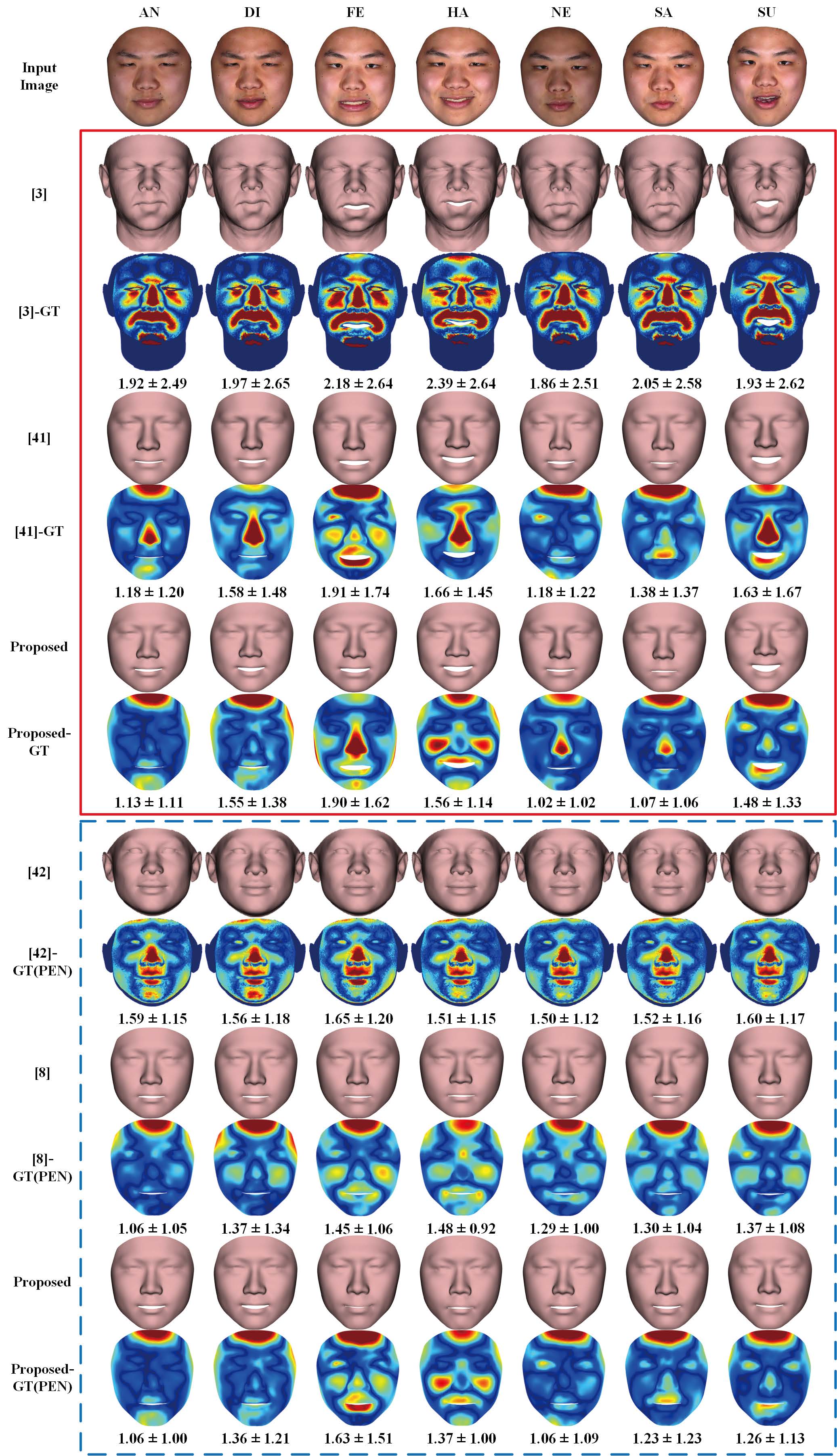}
\caption{Reconstruction results for a BU3DFE subject in seven expressions. The first row shows the input images. The red box shows the reconstructed 3D faces with the same expression as the input, using \cite{liu2015cascaded}, \cite{zhu2015high} and the proposed method. The blue box shows the reconstructed PEN 3D faces by \cite{liu2016joint}, \cite{tran2016regressing} and the proposed method.}
% which go from dark blue to dark red (corresponding to errors between $0$ and $5$). The numbers under each error map represent mean and standard deviation  (in $\%$).}
\label{fig:bu3d_exp_example}
\end{figure*}

\emph{\textbf{Reconstruction accuracy across poses~~}}Table~\ref{tab:bu3d_pose_compare} shows the average MAE of the proposed method under different poses of the input faces. 
For a fair comparison with the counterpart methods, we only compute the reconstruction error of neutral testing images. 
To compute MAE, the reconstructed 3D faces should be first aligned to the ground truth. 
Since the results of~\cite{liu2016joint,liu2015cascaded} and our proposed method already have the same number of vertices as the ground truth, we employ Procrustes alignment for these methods as suggested by~\cite{bas2016fitting}. 
For the results of~\cite{zhu2015high,tran2016regressing}, however, the number of vertices is different from the ground truth. 
Hence, we align them using rigid ICP method as~\cite{tran2016regressing} does. 
It can be seen from Table~\ref{tab:bu3d_pose_compare} that the average MAE of the proposed method (either linear or nonlinear implementation) is lower than that of counterpart methods. 
Moreover, as the pose becomes large, the error of the proposed method does not increase substantially. This proves the effectiveness of the proposed method in handling arbitrary view faces. Figure~\ref{fig:bu3d_pose_example} shows the reconstruction results of one subject.

\emph{\textbf{Reconstruction accuracy across expressions~~}}Figure~\ref{fig:bu3d_exp_error_arxiv} shows the average MAE of the proposed method and \cite{liu2015cascaded,zhu2015high} across expressions, based on their reconstructed 3D faces that have the same pose and expression as the input. 
The proposed method overwhelms its counterpart for all expressions. 
Moreover, as expressions change, the MAE standard deviation of \cite{zhu2015high,liu2015cascaded} are $0.157 mm$ and $0.179 mm$, whereas that of the proposed method is $0.034 mm$ in linear implementation and $0.029 mm$ in nonlinear implementation. 
This proves the superior robustness of the proposed method to expression variations.

Figure~\ref{fig:bu3d_exp_error_eccv} compares the average MAE of the proposed method and \cite{liu2016joint,tran2016regressing} across expressions, based on their reconstructed PEN 3D faces. 
Again, the proposed method shows superiority in both MAE under all expressions and robustness across expressions. 
We believe that such superiority is owing to its explicit modeling of expression deformation. Figure~\ref{fig:bu3d_exp_example} shows the reconstruction results for one subject under seven expressions.

\emph{\textbf{Reconstruction accuracy across races~~}} 
It is well known that people from different races (e.g., Asian and Caucasian) show different characteristics in facial shapes. 
Such other-race effect has been reported in face recognition literature~\cite{phillips2011other}. 
In this experiment, we study the impact of races on 3D face reconstruction using the FRGC v2.0 database~\cite{phillips2005overview}. 
FRGC v2.0 contains 3D faces and images of $466$ subjects with different ethnic groups (Table~\ref{tab:statistics_race}). 
Since these faces have no expression variation, the expression shape component in our proposed model is set to zero. 
We use the method in~\cite{bolkart20153d} to establish dense correspondence of the 3D faces of $5{,}996$ vertices. 
We conduct three  experiments: (i) training with $100$ Asian samples (denoted as \textbf{Setting I}), (ii) training with $100$ Caucasian samples (\textbf{Setting II}), and (iii) training with $100$ Asian and $100$ Caucasian samples (\textbf{Setting III}). 
The testing set contains samples of remaining subjects in FRGC v2.0, including $12$ Asian, $6$ African, $13$ Hispanic, $19$ Caucasian and $16$ Unknown races. % FIXME for set 1 and 2, why not use 200 samples? Then it is more fair comparison with set 3

\begin{table}[!t]
\scriptsize
\renewcommand{\arraystretch}{1.1}
\newcommand{\tabincell}[2]{\begin{tabular}{@{}#1@{}}#2\end{tabular}}
\caption{Number and percentage of subjects of different genders and races in the FRGC v2.0 database.}
\label{tab:statistics_race}
\centering
% Some packages, such as MDW tools, offer better commands for making tables
% than the plain LaTeX2e tabular which is used here.
\begin{tabular}{p{0.8cm} p{0.7cm}<{\centering} p{0.75cm}<{\centering} p{0.9cm}<{\centering} p{1cm}<{\centering} p{1cm}<{\centering} p{0.75cm}<{\centering} }
\hline 
 & Asian & African & Hispanic & Caucasian & Unknown & Total\\
\hline 
\hline 
 Female & \tabincell{c}{55 \\ (\textit{$11.8\%$})} & \tabincell{c}{2 \\ (\textit{$0.4\%$})} & \tabincell{c}{5 \\ (\textit{$1.1\%$})} & \tabincell{c}{134 \\ (\textit{$28.8\%$})} & \tabincell{c}{6 \\ (\textit{$1.3\%$})} & \tabincell{c}{202 \\ (\textit{$43.3\%$})}\\
 \hline 
 Male & \tabincell{c}{57 \\ (\textit{$12.2\%$})} & \tabincell{c}{4 \\ (\textit{$0.9\%$})} & \tabincell{c}{8 \\ (\textit{$1.7\%$})} & \tabincell{c}{185 \\ (\textit{$39.7\%$})} & \tabincell{c}{10 \\ (\textit{$2.1\%$})} & \tabincell{c}{264 \\ (\textit{$56.7\%$})}\\
 \hline 
 Total & \tabincell{c}{112 \\ (\textit{$24.0\%$})} & \tabincell{c}{6 \\ (\textit{$1.3\%$})} & \tabincell{c}{13 \\ (\textit{$2.8\%$})} & \tabincell{c}{319 \\ (\textit{$68.5\%$})} & \tabincell{c}{16 \\ (\textit{$3.4\%$})} & \tabincell{c}{466 \\ (\textit{$100\%$})}\\
\hline 
\end{tabular}
\end{table}

\begin{figure}[t]
\centering
\includegraphics[width=2.7in]{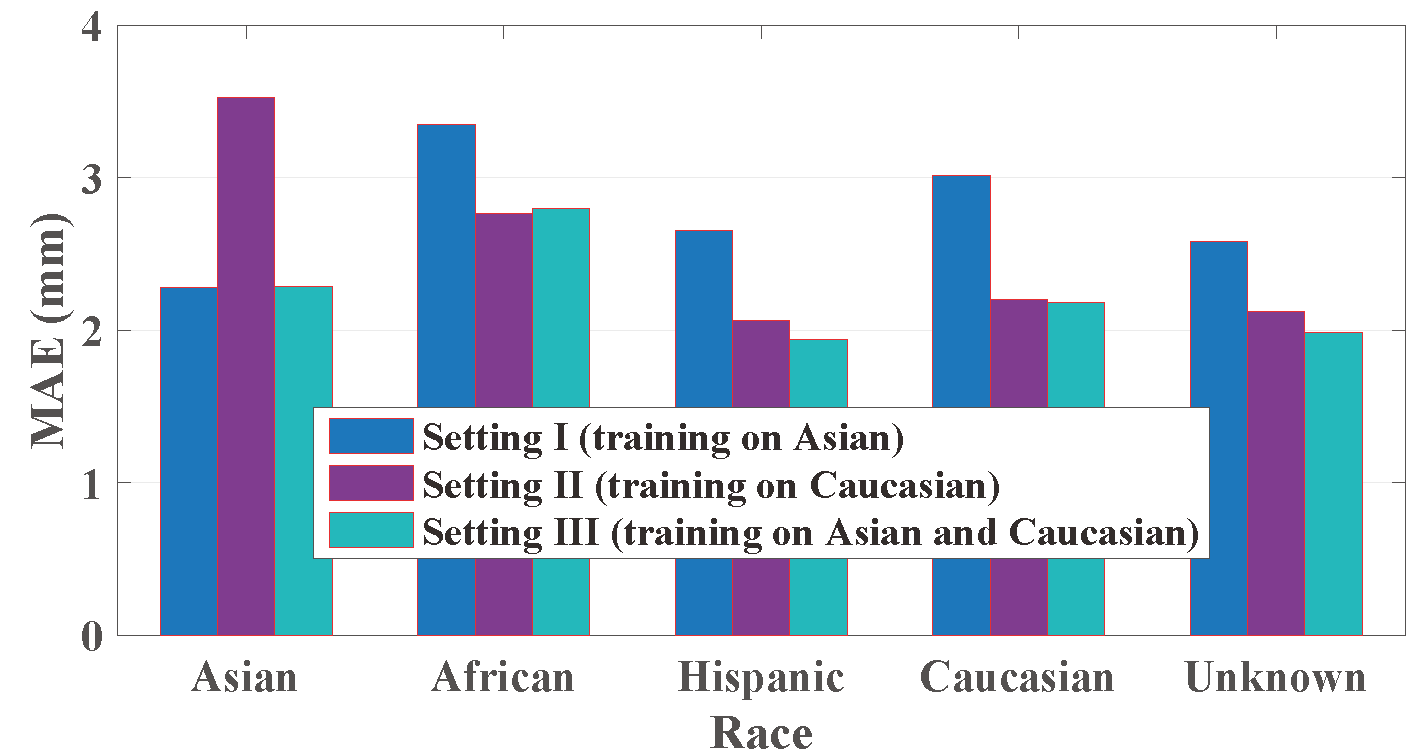}
\caption{3D face reconstruction accuracy (MAE) of the proposed method across different ethnic groups.}
\label{fig:frgc_race_error}
\end{figure}

Figure~\ref{fig:frgc_race_error} compares the 3D face reconstruction accuracy (MAE) across different ethnic groups. 
Not surprisingly, training for one ethnic group can yield higher accuracy on testing of the same ethnic. 
As for the other-race effect, the model trained on Caucasian achieves comparable accuracy on Caucasian and Hispanic, but much worse on the other races (and worst on Asian). 
On the other hand, the model trained on Asian performs much worse on all other races compared to on its own race, and the worst on African. 
These results reveal the variations in the facial shapes of people from different races. 
Further, by combining training data of Asian and Caucasian (Setting III), comparable reconstruction accuracy is achieved for both Asian and Caucasian, which is also comparable to those in Setting I and II. 
This proves the capability of the proposed method in handling the 3D shape variations among {\it all} ethnic groups.

\begin{table*}[!t]
\renewcommand{\arraystretch}{1.1}
\centering 
\caption{The face alignment accuracy (NME) of the proposed method and state-of-the-art methods on AFLW and AFLW2000-3D databases.} 
\begin{tabular}{p{2.7cm} || c c  c | c c || c c c | c c} 
\hline
\multirow{2}{*}{\textbf{Method}}& \multicolumn{5}{c||}{\textbf{AFLW Database (21 points)}} & \multicolumn{5}{c}{\textbf{AFLW2000-3D Database (68 points)}}\\ 

\cline{2-11} % Horizontal line spanning less than the full width of the table - you can add (r) or (l) just before the opening
% curly bracket to shorten the rule on the left or right side
 & $\left [ 0^{\circ}, 30^{\circ} \right )$ & $\left [ 30^{\circ}, 60^{\circ} \right )$ & $\left [ 60^{\circ}, 90^{\circ} \right ]$ & Mean & Std & $\left [ 0^{\circ}, 30^{\circ} \right )$ & $\left [ 30^{\circ}, 60^{\circ} \right )$ & $\left [ 60^{\circ}, 90^{\circ} \right ]$ & Mean & Std\\ % Column names row
\hline % In-table horizontal line
\hline % In-table horizontal line
RCPR~\cite{burgos2013robust}                & $5.43$ & $6.58$  & $11.53$ & $7.85$  & $3.24$  & $4.26$ & $5.96$  & $13.18$ &  $7.80$ & $4.74$\\ 
\hline
ESR~\cite{cao2014face}                  & $5.66$ &  $7.12$ & $11.94$ &  $8.24$ & $3.29$  & $4.60$ &  $6.70$ & $12.67$ &  $7.99$ & $4.19$\\ 
\hline
SDM~\cite{xiong2013supervised}                  & $4.75$ &  $5.55$ &  $9.34$ &  $6.55$ & $2.45$  & $3.67$ &  $4.94$ &  $9.76$ &  $6.12$ & $3.21$\\ 
\hline
3DDFA~\cite{zhu2016CVPR}               & $5.00$ &  $5.06$ &  $6.74$ &  $5.60$ & $0.99 $ & $3.78$ &  $4.54$ &  $7.93$ &  $5.42$ & $2.21$\\ 
3DDFA+SDM~\cite{zhu2016CVPR}             & $4.75$ &  $4.83$ &  $6.38$ &  $5.32$ & $0.92$  & $3.43$ &  $4.24$ &  $7.17$ &  $4.94$ & $1.97$\\
\hline
\hline 
Proposed (Linear)                      & $3.75$ &  $4.33$ &  $5.39$ &  $4.49$ & $\bf{0.83}$   & $3.25$ &  $\bf{3.95}$ &  $6.42$ &  $4.61$ & $1.78$\\
Proposed (Nonlinear)                     & $\bf{3.22}$ &  $\bf{4.13}$ &  $\bf{5.13}$ &  $\bf{4.16}$ & $0.96$  & $\bf{2.72}$ &  $4.06$ &  $\bf{5.81}$ &  $\bf{4.20}$ & $\bf{1.55}$\\
\hline % Bottom horizontal line
\end{tabular}
\label{tab:aflw_compare}
\end{table*}

\begin{figure*}[t]
\begin{center}
\includegraphics[width=7in]{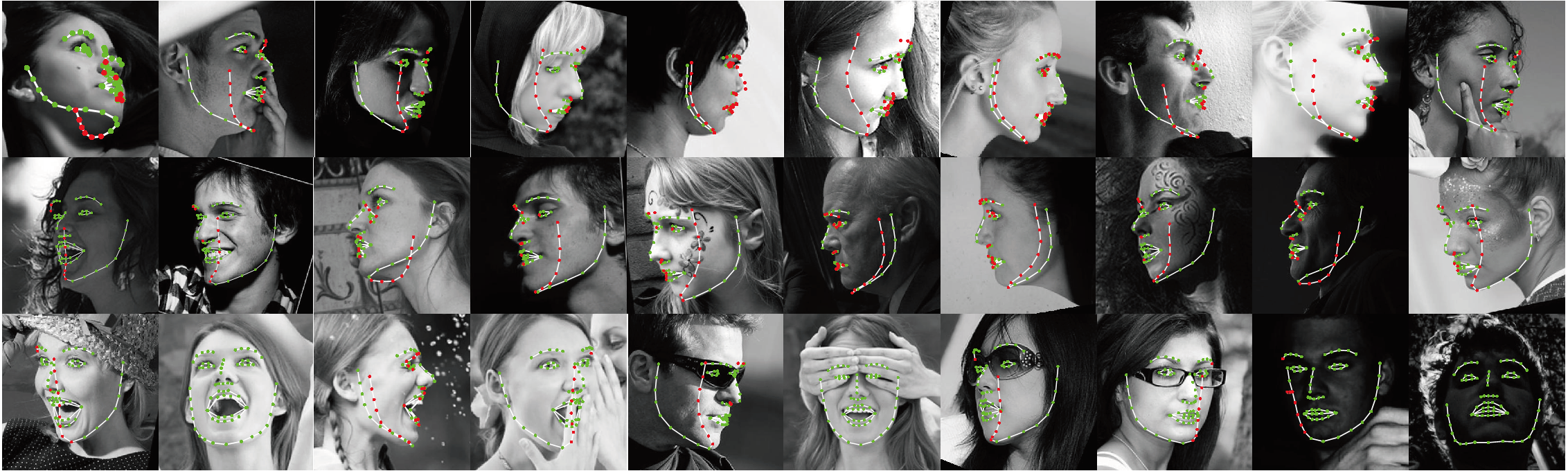}
\end{center}
   \caption{The $68$ landmarks detected by the proposed method for AFLW data. Green/red points denote visible/invisible landmarks.}
\label{fig:land_example}
\end{figure*}

\subsection{Face Alignment Accuracy}
%\subsubsection{Databases}
In evaluating face alignment, several state-of-the-art face alignment methods are considered for comparison to the proposed method, including RCPR~\cite{burgos2013robust}, ESR~\cite{cao2014face}, SDM~\cite{xiong2013supervised}, 3DDFA and 3DDFA+SDM~\cite{zhu2016CVPR}. 
The dataset constructed from 300W-LP is used for training, the AFLW~\cite{K2011Annotated} and AFLW2000-3D~\cite{zhu2016CVPR} are for testing. 
AFLW contains $25{,}993$ in-the-wild faces with large poses (yaw from -$90^{\circ}$ to $90^{\circ}$). 
Each image is annotated with up to $21$ visible landmarks. 
For a fair comparison to \cite{zhu2016CVPR}, we use the same $21{,}080$ samples as our testing set, and divide the testing set into three subsets according to the absolute yaw angle of the testing image: $\left [ 0^{\circ}, 30^{\circ} \right )$, $\left [ 30^{\circ}, 60^{\circ} \right )$ and $\left [ 60^{\circ}, 90^{\circ} \right ]$. 
The resulting three subsets have $11{,}596$, $5{,}457$ and $4{,}027$ samples, respectively. 
AFLW2000-3D contains the ground truth 3D faces and the corresponding $68$ landmarks of the first $2{,}000$ AFLW samples. 
There are $1{,}306$ samples in $\left [ 0^{\circ}, 30^{\circ} \right )$, $462$  in $\left [ 30^{\circ}, 60^{\circ} \right )$ and $232$  in $\left [ 60^{\circ}, 90^{\circ} \right ]$. 
The bounding boxes provided by AFLW are used in the AFLW testing, while the ground truth bounding boxes enclosing all $68$ landmarks are used for the AFLW2000-3D testing.

Normalized Mean Error (NME)~\cite{jourabloo2015pose} is employed to measure the face alignment accuracy. 
It is defined as the mean of the normalized estimation error of visible landmarks for all testing samples:
\begin{equation}
\texttt{NME} = \frac{1}{N_{T}}\sum_{i=1}^{N_{T}}{\left(\frac{1}{d_{i}}\frac{1}{N^{\textbf{v}}_{i}} \sum_{j=1}^{l}{\textbf{v}_{ij}||(\hat{u}_{ij}, \hat{v}_{ij}) - (u^{*}_{ij}, v^{*}_{ij})||} \right)},
\end{equation}
where $d_{i}$ is the square root of the bounding box area of the $i^{\texttt{th}}$ testing sample, $N^{\textbf{v}}_{i}$ is the number of its visible landmarks, $(u^{*}_{ij}, v^{*}_{ij})$ and $(\hat{u}_{ij}, \hat{v}_{ij})$ are, respectively, the ground truth and estimated coordinates of its $j^{\texttt{th}}$ landmark.

Table~\ref{tab:aflw_compare} compares the face alignment accuracy on the AFLW and AFLW2000-3D datasets. 
As can be seen, the proposed method achieves the best accuracy for all poses and on both datasets. 
In order to assess the robustness of different methods to pose variations, we also report their standard deviations of the NME in Table~\ref{tab:aflw_compare}. 
The results again demonstrate the superiority of the proposed method over the counterpart. Figure~\ref{fig:land_example} shows the landmarks detected by the proposed method on some AFLW images.

Moreover, for the proposed method, the nonlinear regression implementation is better than the linear one. 
CNN feature is more powerful and robust than the handcrafted SIFT feature for the face alignment task. In contrast, in the experiments of 3D face reconstruction on BU3DFE database (see Section 5.1), the reconstruction error of linear regressors is lower than that of nonlinear regressors. This might be because MLP-based nonlinear regressors for 3D face reconstruction need more training samples.

\subsection{Face Recognition}
While there are many recent face alignment and reconstruction works~\cite{guler2016densereg, tuzel2016robust, peng2016recurrent, booth20163d, richardson2016learning}, few works take one step further to evaluate the contribution of alignment or reconstruction to subsequent tasks, such as face recognition. 
In contrast, we quantitatively evaluate the contribution of the reconstructed pose-expression-normalized (PEN) 3D faces to face recognition by directly matching 3D to 3D shape and fusing it with conventional 2D face recognition. 
Refer to Sec.~\ref{sec:facerecognition} for details of the PEN 3D faces enhanced face recognition method. %The linear implementation of the proposed method is employed in the following experiments.

\begin{table}[t]
\centering 
\renewcommand{\arraystretch}{1}
\scriptsize
\caption{Recognition accuracy (\%) in the first experiment on Multi-PIE by the four state-of-the-art DL-based face matchers before (indicated by suffix ``2D'') and after (indicated by suffix ``Fusion'') our 3D enhancement.} % Avg.~is the average accuracy. } 
\begin{tabular*}{8.8cm}{p{2.6cm} p{0.41cm}<{\centering} p{0.41cm}<{\centering} p{0.41cm}<{\centering} p{0.41cm}<{\centering} p{0.41cm}<{\centering} p{0.41cm}<{\centering} p{0.41cm}<{\centering} p{0.41cm}<{\centering}}
\toprule
%& \multicolumn{7}{c}{\textbf{Rotation angle}} \\ % Amalgamating several columns into one cell is done using the \multicolumn command as seen on this line
%\cmidrule(l){2-8} % Horizontal line spanning less than the full width of the table - you can add (r) or (l) just before the opening curly bracket to shorten the rule on the left or right side
\multirow{2}{*}\textbf{Method} & $\pm90^{\circ}$ & $\pm75^{\circ}$ & $\pm60^{\circ}$ & $\pm45^{\circ}$ & $\pm30^{\circ}$ & $\pm15^{\circ}$ &\textbf{Avg.} \\ % Column names row
\midrule % In-table horizontal line
\midrule % In-table horizontal line
% Zhu et al. \cite{zhu2013deep} & \text{-}  & \text{-}  & $45.9\%$ & $64.1\%$ & $80.7\%$ & $90.7\%$ & $94.3\%$ & $72.9\%$ & \text{-} \\
% Zhu et al. \cite{zhu2014multi} & \text{-}  & \text{-}  & $60.1\%$ & $72.9\%$ & $83.7\%$ & $92.8\%$ & $95.7\%$ & $79.3\%$ & \text{-} \\
% Yim et al. \cite{yim2015rotating} & \text{-}  & \text{-}  & $61.9\%$ & $79.9\%$ & $88.5\%$ & $95.0\%$ & $99.5\%$ & $83.3\%$ & \text{-} \\
% DR-GAN \cite{LuanPose2017}  & \text{-}  & \text{-}  & $83.2\%$ & $86.2\%$ & $90.1\%$ & $94.0\%$ & $97.0\%$ & $89.2\%$ & \text{-} \\
 VGG-2D  & $36.2$ & $66.9$ & $83.5$ & $93.8$ & $97.7$ & $98.6$ & $79.5$\\ 
 LightenedCNN-2D & $7.50$ & $31.5$ & $78.6$ & $96.3$ & $99.1$ & $99.8$ & $68.8$\\ 
 CenterLoss-2D & $48.2$ & $72.7$ & $92.6$ & $98.8$ & $99.6$ & $99.7$ & $85.3$\\ 
LDF-Net-2D & $65.3$ & $ 86.2$ & $93.7$ & $98.4$ & $98.9$ & $98.6$ & $90.2$\\ 
 \midrule
  ICP-3D & $31.8$ & $30.6$ & $34.3$ & $32.8$ & $34.7$ & $44.3$ & $33.0$ \\ 
  \midrule
%  $w_{2d}=0.5$,  $w_{3d}=0.5$\\ 
   VGG-Fusion& $52.6$ & $75.2$ & $90.5$ & $96.8$ & $98.5$ & $99.4$ & $85.5$ \\ 
  LightenedCNN-Fusion & $23.6$ & $45.3$ & $84.6$ & $97.6$ & $99.6$ & $99.9$ & $75.1$\\ 
  CenterLoss-Fusion & $63.7$ & $76.7$ & $92.5$ & $97.8$ & $98.4$ & $98.7$ &  $88.0$\\ 
LDF-Net-Fusion  & $70.4$ & $87.6$ & $93.4$ & $98.1$ & $97.9$ & $97.7$ & $90.9$\\ 
\bottomrule % Bottom horizontal line
\end{tabular*}
\label{tab:multipie_setting1}
\end{table}

\begin{table*}[t]
\footnotesize
\renewcommand{\arraystretch}{1}
\caption{Recognition accuracy (\%) of the CenterLoss matcher in the second experiment on Multi-PIE. The results shown in brackets are obtained by using the original CenterLoss matcher without enhancement by our reconstructed 3D faces.}
\label{tab:multipie_setting2}  
\centering
% Some packages, such as MDW tools, offer better commands for making tables
% than the plain LaTeX2e tabular which is used here.
\begin{tabular}{|c |c |c |c |c |c |c|}
\hline 
Pose $\backslash$ Expression & Smile & Surprise & Squint & Disgust & Scream & \textbf{Avg.}\\
\hline 
\hline 
$\pm90^{\circ}$ & $51.4 (36.9)$ & $46.1 (35.7)$ & $58.8 (38.7)$ & $42.0 (24.9)$ & $63.6 (52.4)$ & $52.4 (37.7)$\\
 \hline 
$\pm75^{\circ}$ & $73.1 (67.0)$ & $56.6 (53.0)$ & $72.6 (67.8)$ & $52.5 (43.4)$ & $75.1 (71.6)$ & $66.0 (60.4)$\\
 \hline 
$\pm60^{\circ}$ & $88.6 (89.8)$ & $80.2 (80.7)$ & $91.6 (88.2)$ & $74.6 (69.8)$ & $91.8 (92.7)$ & $85.4 (84.2)$\\
 \hline 
$\pm45^{\circ}$ & $95.9 (97.6)$  & $89.4 (95.1)$  & $95.6 (97.8)$ & $86.7 (83.5)$  & $97.3 (98.7)$  & $93.0 (94.5)$ \\
 \hline 
$\pm30^{\circ}$ & $97.8 (99.1)$  & $93.1 (97.0)$  & $96.8 (99.3)$  & $90.4 (91.5)$  & $98.5 (99.8)$ & $95.3 (97.3)$ \\
 \hline 
$\pm15^{\circ}$ & $98.5 (99.6)$  & $95.6 (97.3)$  & $97.5 (100)$   & $92.6 (93.5)$ & $98.1 (99.2)$  & $96.5 (97.9)$ \\
 \hline 
 \textbf{Avg.} & $84.2 (81.7)$ & $76.8 (76.5)$  & $85.5 (82.0)$  & $73.1 (67.8)$ & $87.4 (85.7)$  & $81.4 (78.7)$ \\
%$ 0^{\circ}$ & $99.4 (99.6)$  & $96.1 (98.0)$ & $98.0 (99.5)$ & $95.2 (93.9)$  & $98.7 (99.2)$ & $97.5 (98.0)$ \\
% \hline 
%\textbf{Avg.} & $86.4 (84.2)$ & $79.6 (79.6)$  & $87.3 (84.5)$  & $76.3 (71.4)$ & $89.0 (87.6)$  & $83.7 (81.5)$ \\
\hline 
\end{tabular}
\end{table*}

In this evaluation,  we employ the linear implementation, and use the BU3DFE ($13,300$ images of $100$ subjects; refer to Sec.~\ref{sec:train_data_preparation}) and MICC~\cite{bagdanov2011florence} databases as training data, the CMU Multi-PIE database~\cite{gross2010multi} and the Celebrities in Frontal-Profile (CFP) database~\cite{sengupta2016frontal} as test data. 
MICC contains 3D face scans and video clips (indoor, outdoor and cooperative head rotations environments) of $53$ subjects. 
We randomly select faces with different poses from the cooperative environment videos, resulting in $11{,}788$ images of $53$ subjects and their corresponding neutral 3D face shapes (whose expression shape components are thus set to zero). The 3D faces are processed by the method in~\cite{bolkart20153d} to establish dense correspondence with $n=5{,}996$ vertices.

\subsubsection{Face Identification on Multi-PIE Database}

CMU Multi-PIE is a widely used benchmark database for face recognition, %under pose, illumination and expression variations. 
with faces of $337$ subjects collected under various views, expressions and lighting conditions. 
Here, we consider pose and expression variations, and conduct two experiments. 
In the first experiment, following the setting of~\cite{zhu2014multi, zhu2015high}, probe images consist of the images of all $337$ subjects at $12$ poses ($\pm90^{\circ}$, $\pm75^{\circ}$, $\pm60^{\circ}$, $\pm45^{\circ}$, $\pm30^{\circ}$, $\pm15^{\circ}$) with neutral expression and frontal illumination. 
In the second experiment, instead of neutral expression, all images with smile, surprise, squint, disgust and scream expressions at the $12$ poses and under frontal illumination are the probe images. 
This protocol is an extended version of \cite{chu20143d,zhu2015high} by adding large-pose images ($\pm60^{\circ}$, $\pm75^{\circ}$, $\pm90^{\circ}$). 
In both experiments, the frontal images captured in the first session are the gallery. 
And four state-of-the-art deep learning based (DL-based) face matchers are used as baselins, i.e., VGG~\cite{parkhi2015deep}, Lightened CNN~\cite{wu2015light}, CenterLoss~\cite{wen2016discriminative} and LDF-Net~\cite{hu2017FG}. 
The first three matchers are publicly available. 
We evaluate them with all $337$ subjects in Multi-PIE. 
The last matcher, LDF-Net, is a latest one specially designed for pose-invariant face recognition. 
It uses the first $229$ subjects for training and the remaining $108$ subjects for testing. 
Since it is not publicly available, we request the match scores from the authors, and fuse our 3D shape match scores with theirs. 
Note that given the good performance of LDF-Net, we assign a higher weight (i.e., $0.7$) to it, whereas the weights for all the other three baseline matchers are set to $0.5$.

Table~\ref{tab:multipie_setting1} reports the rank-1  accuracy of the baseline face matchers in the first experiment, 
%According to the results in Table \ref{tab:multipie_setting1}, 
where the baseline matchers are all further improved by our proposed method. 
Specifically, VGG and Lightened CNN are consistently improved across different poses when fused with 3D, while CenterLoss gains substantial improvement at large poses ($15.5\%$ at $\pm90^{\circ}$ and $4.0\%$ at $\pm75^{\circ}$). 
Even for the latest LDF-Net, the recognition accuracy is improved by $5.1\%$ at $\pm90^{\circ}$ and $1.4\%$ at $\pm75^{\circ}$. 
For all the baseline matchers, the larger the yaw angle is, the more evident the accuracy improvement. 
Table~\ref{tab:multipie_setting1} also gives the recognition accuracy of using only the reconstructed 3D faces, at the row headed by ``ICP-3D''. 
Although its average accuracy is much worse compared with its 2D counterparts, it fluctuates more gently as probe faces rotate from frontal to profile. 
These results prove the effectiveness of the proposed method in dealing with pose variations, as well as in reconstructing individual 3D faces with discriminative details that are complementary to 2D face recognition.

Given its best performance among three publicly available baseline matchers, we employ the CenterLoss matcher in the second experiment. 
The results are shown in Table~\ref{tab:multipie_setting2}. 
As can be seen, the compound impact of pose and expression variations makes the face recognition more challenging, resulting in obviously lower accuracy compared with those in Table~\ref{tab:multipie_setting1}. 
Yet, our proposed method still improves the overall accuracy of the baseline, especially for probe faces of large pose or disgust expression. 
We believe that such performance gain in recognizing non-frontal and expressive faces is owing to the capability of the proposed method in providing complementary pose-and-expression-invariant discriminative features in the 3D face shape space.

\subsubsection{Face Verification on CFP Database}

We further evaluate our reconstructed PEN 3D faces on a more challenging unconstrained face recognition setting by using the CFP database, which has $500$ subjects each with $10$ frontal and $4$ profile images. 
The evaluation includes frontal-frontal (FF) and frontal-profile (FP) face verification, each having $10$ folders with $350$ same-person and $350$ different-person pairs. 
Table~\ref{tab:cfp_setting} reports the average results with standard deviations in terms of Accuracy, Equal Error Rate (EER), and Area Under the Curve (AUC). 

\begin{figure}[t]
\begin{center}
\includegraphics[width=2.8in]{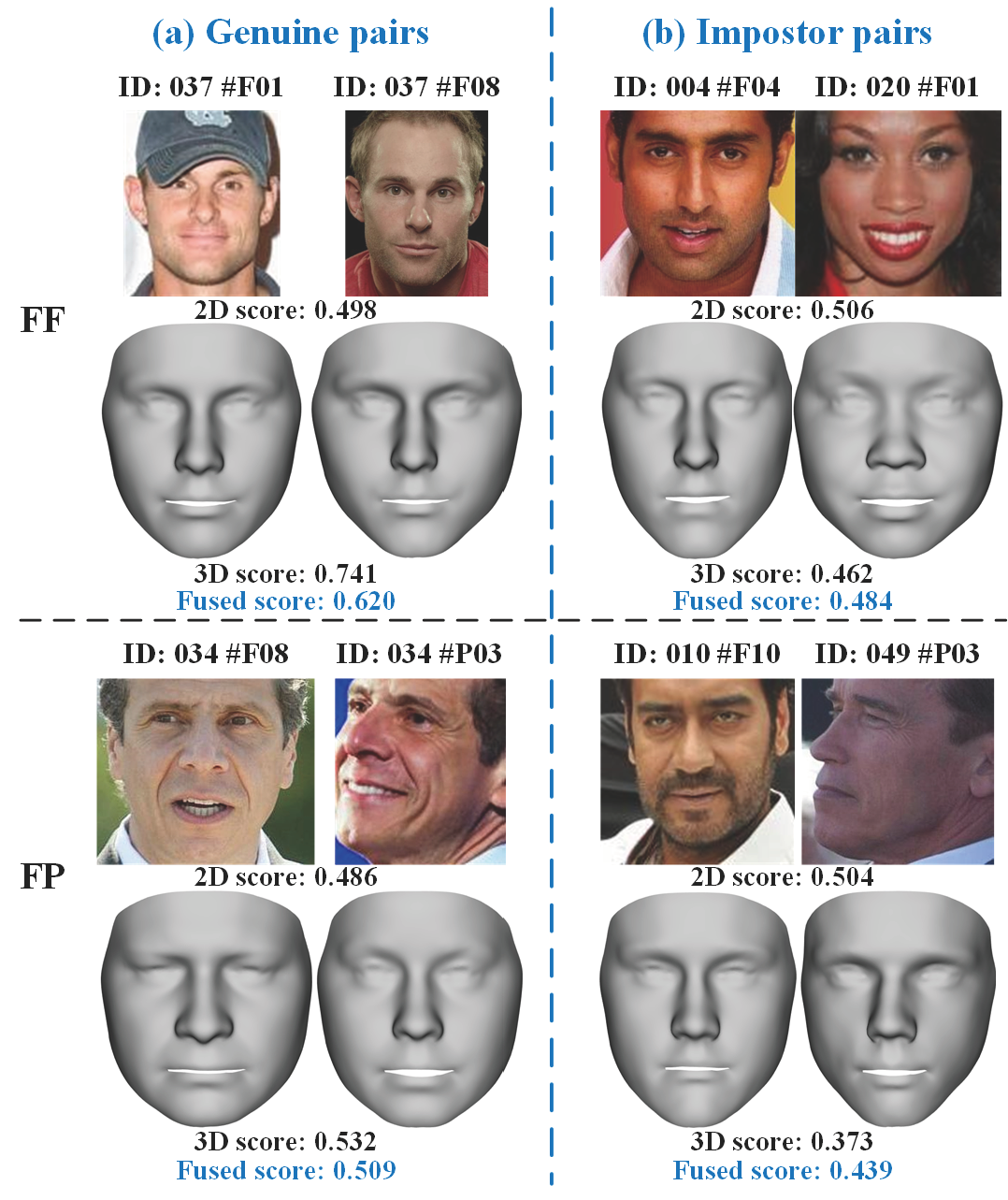}
\end{center}
   \caption{Example (a) genuine pairs and (b) imposter pairs in CFP and corresponding PEN 3D faces, for which the CenterLoss method fails, whereas its fusion with our proposed method succeeds. Note that the operational threshold in our experiments is empirically set to $0.502$.}
\label{fig:recog_example}
\end{figure}

Given its best performance on Multi-PIE database, we employ the CenterLoss matcher in this experiment. 
We also report the recognition accuracy of reconstructed PEN 3D faces (see ``ICP-3D''). 
Although its average accuracy is much worse compared with the baseline, it further improves the performance of CenterLoss in both frontal-frontal (FF) and frontal-profile (FP) face verification. 
These results prove the effectiveness of the proposed method in dealing with pose variations, as well as the ability in providing complementary discriminative features in unconstrained environment. Figure~\ref{fig:recog_example} shows some example genuine and imposter pairs in CFP, which are incorrectly recognized by CenterLoss, but correctly recognized by fusion of CenterLoss and our proposed method.

\begin{table}[t]
\centering 
\renewcommand{\arraystretch}{1}
\scriptsize
\caption{Verification accuracy on CFP by the CenterLoss face matchers before (indicated by suffix ``2D'') and after (indicated by suffix ``Fusion'') the enhancement by our proposed method.} 
\begin{tabular*}{8.9cm}{l l c c c }
\toprule
%& \multicolumn{7}{c}{\textbf{Rotation angle}} \\ % Amalgamating several columns into one cell is done using the \multicolumn command as seen on this line
%\cmidrule(l){2-8} % Horizontal line spanning less than the full width of the table - you can add (r) or (l) just before the opening curly bracket to shorten the rule on the left or right side
\multicolumn{2}{c}{\textbf{Method}}  & CenterLoss-2D & ICP-3D  & CenterLoss-Fusion\\ % Column names row
\midrule
\midrule
\multirow{3}{*}{FF} & Accuracy (\%) & $86.43 \pm3.10$ & $74.83 \pm3.85$ & $89.21 \pm2.88$ \\
                    & EER      (\%) & $14.20 \pm3.58$ & $27.65 \pm3.80$ & $11.37 \pm2.94$ \\
                    & AUC      (\%) & $93.38 \pm2.18$ & $78.41 \pm4.11$ & $94.24 \pm2.67$ \\
\midrule
\multirow{3}{*}{FP} & Accuracy (\%) & $69.27 \pm2.33$ & $65.74 \pm2.47$ & $72.99 \pm1.90$ \\                
                    & EER      (\%) & $31.63 \pm2.36$ & $36.26 \pm2.76$ & $27.91 \pm2.06$ \\
                    & AUC      (\%) & $74.61 \pm2.54$ & $69.02 \pm3.69$ & $78.64 \pm2.43$ \\
                  
\bottomrule % Bottom horizontal line
\end{tabular*}
\label{tab:cfp_setting}
\end{table}

\subsection{Convergence}
\begin{figure}[t]
\centering
\subfigure[]{\includegraphics[width=1.72in]{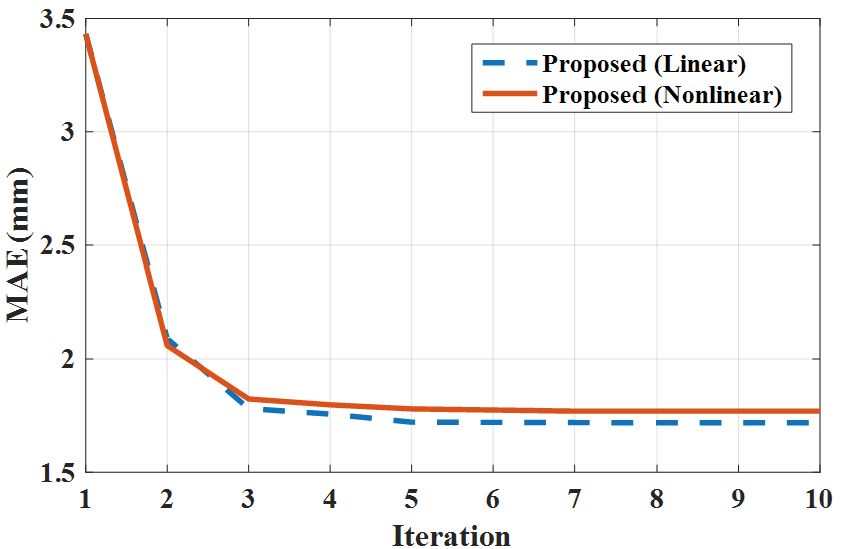}}
\subfigure[]{\includegraphics[width=1.72in]{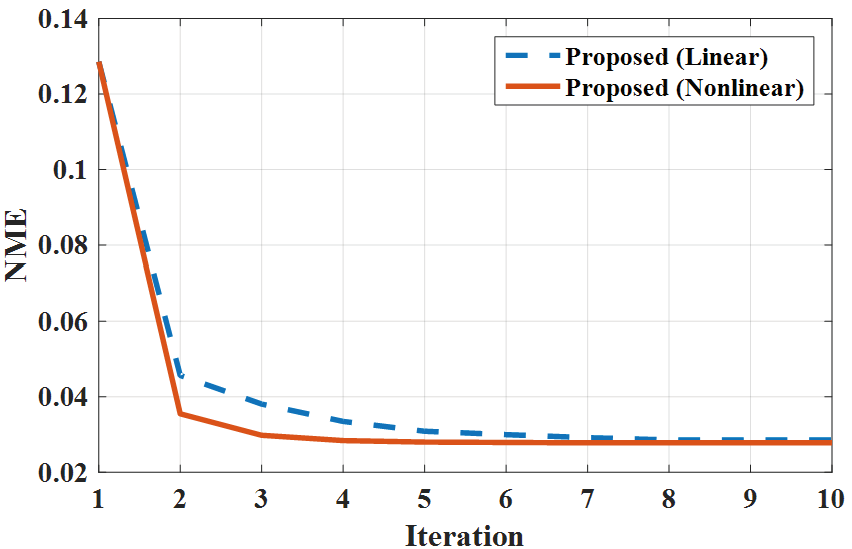}}
\caption{ (a) and (b) show the reconstruction errors (MAE) and alignment errors (NME) during the training of proposed method as iteration proceeds, when trained on the BU3DFE database.} %FIXME the vertical axis of a and c should be the same? b and d should be the same? change caption accordingly. Also, why the vertical axis of a/c have DIFFERENT ranges? Are they optimizing the same objective? If they are the same, we could combine a/c to one figure, b/d to one figure
\label{fig:iteration}
\end{figure}

The proposed method has two alternate optimization processes, one in 2D space for face alignment and the other in 3D space for 3D shape reconstruction. 
We experimentally investigate the convergence of these two processes when training the proposed linear and nonlinear implementations on the BU3DFE database. 
We conduct ten-fold cross-validation experiments, and compute the average errors over the training data through ten iterations. 
As shown in Fig.~\ref{fig:iteration}, the training errors converge in about five iterations in the linear implementation, while in the nonlinear implementation the training errors converge fast after two to three iterations. 
Hence, we set the number of iterations as $K = 5$ and $K=3$ in the linear and nonlinear implementations, respectively. 

\subsection{Computational Complexity}
 According to our experiments on a PC with i7-4790 CPU and $32$ GB memory, the linear implementation of the proposed method runs at $\sim 26$ FPS, and the nonliner implementation runs at $\sim 52$ FPS with a NVIDIA GeForce GTX $1080$. 
This indicates that the proposed method can detect landmarks and reconstruct 3D faces in {\it real-time}. 
We also report the efficiency of individual steps in Table~\ref{tab:run_time}, and comparison with existing methods in Table~\ref{tab:run_time_comparsion}.
 
\begin{table}[t]
\footnotesize
\renewcommand{\arraystretch}{1}
\newcommand{\tabincell}[2]{\begin{tabular}{@{}#1@{}}#2\end{tabular}}
\caption{The time efficiency (in milliseconds or $ms$) of the proposed method.}
\label{tab:run_time}
\centering
\begin{tabular}{l c c c c}
\hline
Step & \tabincell{c}{Updating \\ landmarks} & \tabincell{c}{Updating \\ shape} &  \tabincell{c}{Refining \\ landmarks} & Total \\
\hline
\hline
Linear ($ms$) & $14.93$ & $15.38$ & $8.57$ & $38.88$\\
Nonlinear ($ms$) & $10.22$ & $0.04$ & $9.28$ & $19.32$\\   
\hline
\end{tabular}
\end{table}

\begin{table}[t!]
\footnotesize
\renewcommand{\arraystretch}{1}
\newcommand{\tabincell}[2]{\begin{tabular}{@{}#1@{}}#2\end{tabular}}
\caption{Efficiency comparison of different reconstruction methods. For the methods of~\cite{zhu2015high,tran2016regressing,liu2015cascaded}, although stand-alone landmark detection is required, it is not included in the reported times.}
\label{tab:run_time_comparsion}
\centering
\begin{tabular}{l c c c c c c}
\hline
Method & \cite{zhu2015high} & \cite{tran2016regressing} & \cite{liu2015cascaded} & \cite{liu2016joint} & \tabincell{c}{Proposed \\ (Linear)} & \tabincell{c}{Proposed \\ (Nonlinear)}\\
\hline
\hline
Time ($ms$) & $56.3$ & $88.0$ & $12.6$ & $32.8$ & $38.9$ & $19.3$\\ 
\hline
\end{tabular}
\end{table}

\section{Conclusion}
\label{sec:con}
In this paper, we present a novel regression based method for joint face alignment and 3D face reconstruction from single 2D images of arbitrary poses and expressions. It utilizes landmarks on a 2D face image as clues for reconstructing 3D shapes, and uses the reconstructed 3D shapes to refine landmarks. By alternately applying cascaded landmark regressors and 3D shape regressors, the proposed method can effectively accomplish the two tasks simultaneously in real-time. Unlike existing 3D face reconstruction methods, the proposed method does not require additional face alignment methods, but can fully automatically reconstruct both pose-and-expression-normalized and expressive 3D faces from a single face image of arbitrary poses and expressions. Compared with existing face alignment methods, the proposed method can effectively handle invisible and expression-deformed landmarks with the assistance of 3D face models. Extensive experiments with comparisons to state-of-the-art methods demonstrate the effectiveness and superiority of the proposed method in both face alignment and 3D face reconstruction, and in facilitating cross-view and cross-expression face recognition as well.

%\ifCLASSOPTIONcompsoc
%  % The Computer Society usually uses the plural form
%\else
%  % regular IEEE prefers the singular form
%\fi

\section*{Acknowledgments}

The authors would like to thank the authors of LDF-Net for sharing us with the match scores of LDF-Net on Multi-PIE. This work is supported by the National Key Research and Development Program of China (2017YFB0802300), the National Natural Science Foundation of China (61773270), and the National Key Scientific Instrument and Equipment Development Projects of China (2013YQ49087904). 


\ifCLASSOPTIONcaptionsoff
  \newpage
\fi

\bibliographystyle{IEEEtran}
\bibliography{IEEEabrv,egbib2}


\begin{IEEEbiography}[{\includegraphics[width=1in,height=1.25in,clip,keepaspectratio]{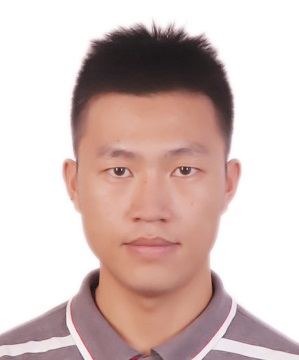}}]{Feng Liu}
is currently a post-doc researcher in the Computer Vision lab at Michigan State University. He received the Ph.D. degree in Computer Science from Sichuan University in 2018. His main research interests are computer vision and pattern recognition, specifically for face modeling, 2D and 3D face recognition. He is a member of the IEEE.
\end{IEEEbiography}

\begin{IEEEbiography}[{\includegraphics[width=1in,height=1.25in,clip,keepaspectratio]{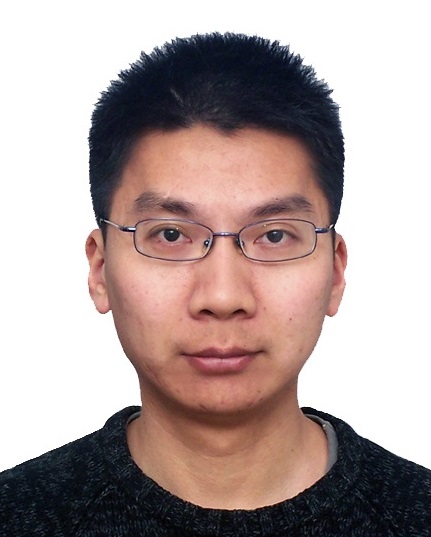}}]{Qijun Zhao}
obtained B.Sc. and M.Sc. degrees both from Shanghai Jiao Tong University, and Ph.D. degree from the Hong Kong Polytechnic University. He worked as a post-doc researcher in the Pattern Recognition and Image Processing lab at Michigan State University from 2010 to 2012. He is currently an associate professor in College of Computer Science at Sichuan University. His research interests lie in biometrics, particularly, face perception, fingerprint recognition, and affective computing, with applications to forensics, intelligent video surveillance, mobile security, healthcare, and human-computer interactions. Dr. Zhao has published more than 60 papers in academic journals and conferences, and participated in many research projects either as principal investigators or as primary researchers. He is a program committee co-chair of CCBR 2016 and ISBA 2018, and an area co-chair of BTAS 2018.
\end{IEEEbiography}

\begin{IEEEbiography}[{\includegraphics[width=1in,height=1.25in,clip,keepaspectratio]{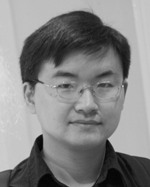}}]{Xiaoming Liu} is an Associate Professor at the Department of Computer Science and Engineering of Michigan State University. He received the Ph.D. degree in Electrical and Computer Engineering from Carnegie Mellon University in 2004. His research interests include computer vision, patter recognition, biometrics and machine learning. He is the recipient of 2018 Withrow Distinguished Scholar Award from Michigan State University. As a co-author, he is a recipient of Best Industry Related Paper Award runner-up at ICPR 2014, Best Student Paper Award at WACV 2012 and 2014, and Best Poster Award at BMVC 2015. He has been an Area Chair for numerous conferences, including FG, ICPR, WACV, ICIP, and CVPR. He is a Co-Program Chair of BTAS 2018 and WACV 2018 conferences. He is an Associate Editor of Neurocomputing journal. He is a guest editor for IJCV Special Issue on Deep Learning for Face Analysis, and ACM TOMM Special Issue on Face Analysis for Applications. He has authored more than 100 scientific publications, and has filed 26 U.S. patents. 
\end{IEEEbiography}

\begin{IEEEbiography}[{\includegraphics[width=1in,height=1.25in,clip,keepaspectratio]{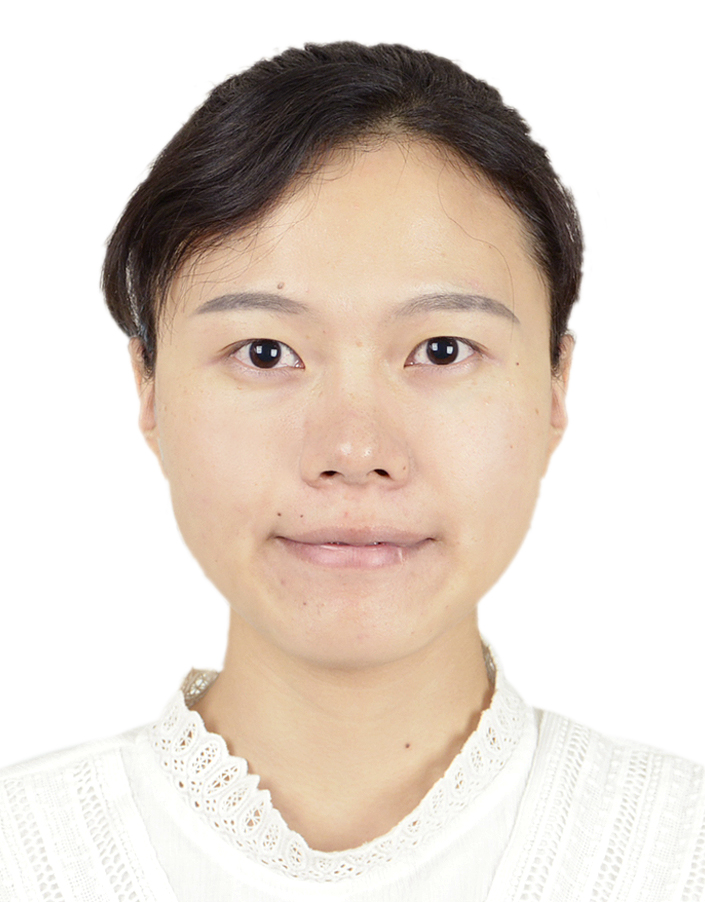}}]{Dan Zeng}
 is currently a post-doc researcher in the DMB group, University of Twente. She received the B.Sc. and Ph.D. degrees from Sichuan University in 2013 and 2018. Since 2012, she participated in '3+2+3' successive graduate, postgraduate and doctoral program of Sichuan University. Her main research area is computer vision and biometrics, specifically for challenges in face recognition.
\end{IEEEbiography}

%\begin{IEEEbiographynophoto}{XXX XXX}

%\newpage

%\begin{IEEEbiographynophoto}{XXX XXX}


%\vfill


\end{document}